\documentclass[11pt]{article}

\usepackage[final]{acl}

\usepackage{times}
\usepackage{latexsym}

\usepackage[T1]{fontenc}
\usepackage[utf8]{inputenc}

\usepackage{microtype}

\usepackage{inconsolata}

\usepackage{graphicx}

\usepackage{multirow}
\usepackage{color}
\usepackage{amsmath}
\usepackage{xcolor}
\usepackage{listings, xcolor}
\usepackage{colortbl}
\usepackage{xspace}
\usepackage{enumitem}
\usepackage{booktabs}
\usepackage{bbding}
\usepackage{pifont}
\usepackage{graphicx}
\usepackage{subcaption}
\usepackage{pifont}
\usepackage{wrapfig}
\usepackage{tcolorbox}
\usepackage{listings}
\tcbuselibrary{skins}
\usepackage{tabularx}
\usepackage{makecell}
\usepackage{pifont}
\usepackage{float}
\usepackage{enumitem}
\usepackage{makecell}
\usepackage{amsfonts}
\usepackage{multirow}
\usepackage{color}
\usepackage{amsmath}
\usepackage{xcolor}
\usepackage{listings, xcolor}
\usepackage{colortbl}
\usepackage{xspace}
\usepackage{enumitem}
\usepackage{booktabs}
\usepackage{bbding}
\usepackage{pifont}
\usepackage{graphicx}
\usepackage{enumitem}
\usepackage{booktabs}
\usepackage{bbding}
\usepackage{pifont}
\usepackage{algorithm}

\newcommand{\ourmodel}{{\texttt{\textbf{Align-GRAG}}}\xspace}

\title{Align-GRAG: Anchor and Rationale Guided Dual Alignment for
Graph Retrieval-Augmented Generation}

\author{Derong Xu*\textsuperscript{1,2}, Pengyue Jia*\textsuperscript{2}, Xiaopeng Li\textsuperscript{2}, Yingyi Zhang\textsuperscript{2}, Maolin Wang\textsuperscript{2}, Qidong Liu\textsuperscript{2},\\
{\bf Xiangyu Zhao\textsuperscript{2}, Yichao Wang\textsuperscript{3}, Huifeng Guo\textsuperscript{3}, Ruiming Tang\textsuperscript{3}, Enhong Chen\textsuperscript{1}, Tong Xu\textsuperscript{1}}\\
\textsuperscript{1}University of Science and Technology of China,\\
\textsuperscript{2}City University of Hong Kong, 
\textsuperscript{3}Noah's Ark Lab, Huawei\\
\textsuperscript{*}Equal contribution.
}

\begin{document}
\maketitle
\begin{abstract}
Despite the strong abilities, large language models (LLMs) still suffer from hallucinations and reliance on outdated knowledge, raising concerns in knowledge-intensive tasks. Graph-based retrieval-augmented generation (GRAG) enriches LLMs with knowledge by retrieving graphs leveraging relational evidence, but it faces two challenges: structure-coupled irrelevant knowledge introduced by neighbor expansion and structure-reasoning discrepancy between graph embeddings and LLM semantics. We propose \ourmodel, an anchor-and-rationale guided refinement framework to address these challenges. It prompts an LLM to extract anchors and rationale chains, which provide intermediate supervision for \textbf{(1) node-level alignment} that identifies critical nodes and prunes noisy evidence, and \textbf{(2) graph-level alignment} that bridges graph and language semantic spaces via contrastive learning. Extensive experiments on commonsense reasoning, scene graph understanding, and knowledge graph reasoning demonstrate consistent gains over 18 strong baselines, validating the effectiveness of \ourmodel for improving graph-grounded generation. The code can be found in \url{https://anonymous.4open.science/r/Align-GRAG-F3D8/}.
\end{abstract}

\section{Introduction}
Recent progress in large language models (LLMs) has highlighted their strong abilities in understanding, reasoning \cite{zhao2023survey,xu2024large}, and knowledge-intensive tasks \cite{zhu2023large,alkhamissi2022review,sui-etal-2025-fidelis}. Despite these advances, LLMs still raise reliability concerns, including hallucinations (producing false or misleading information) \cite{huang2023survey,xu2024hallucination} and reliance on outdated data \cite{fang-etal-2025-karpa}. These issues are especially concerning in high-stakes fields such as healthcare \cite{he2023survey} and law \cite{lai2024large}.

Retrieval-augmented generation (RAG) addresses these concerns by retrieving query-relevant evidence from external databases and grounding generation in a verifiable context, improving factuality and robustness \cite{fan2024survey,zhou-etal-2025-reflection}. Nevertheless, in real-world scenarios, RAG systems often divide long context into independent chunks, overlooking the deeper connections between fragments and lacking a global perspective \cite{lightrag}. Moreover, many knowledge sources are inherently graph-structured, including recommendation systems \cite{wang-etal-2025-knowledge-graph}, the Web, and knowledge graphs \cite{pan2024unifying}. Motivated by this, graph-based RAG (GRAG) \cite{edge2024localgraphrag,G-retriever,gnp,tog} extends RAG by retrieving subgraphs rather than isolated passages \cite{G-retriever}, preserving correlations among multiple text units.

\begin{figure}[t]
\centering
\includegraphics[width=0.85\linewidth]{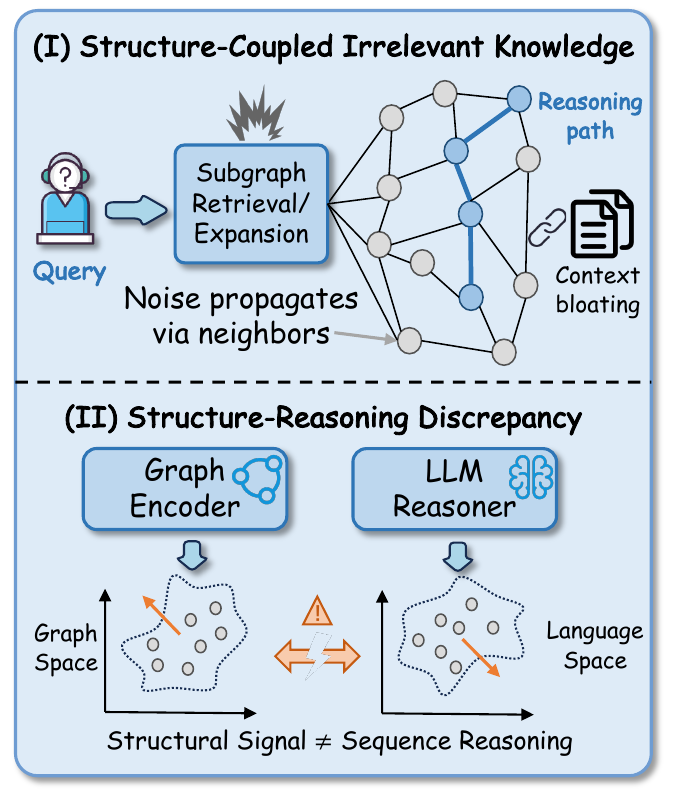}
\caption{The challenges in graph RAG scenarios.}
\label{fig:intro}
\end{figure}

However, integrating RAG with graphs is non-trivial, raising two GRAG-specific challenges, as shown in Figure~\ref{fig:intro}.
\textbf{(I) Structure-Coupled Irrelevant Knowledge.}
To ensure multi-hop coverage, GRAG typically does expansions with retrieved interconnected neighbors. While necessary, this process introduces ``structure-coupled'' noise--irrelevant nodes that are topologically close to the target but semantically unhelpful for the specific reasoning chain. Crucially, determining the utility of these nodes requires a holistic view of the reasoning path rather than assessing local query similarity. Existing post-retrieval rerankers \cite{bge,gte,jia2024bridging}, designed for independent text segments, fail in this context: by treating nodes or triples as isolated units, they sever the structural dependencies essential for judging reasoning-dependent relevance. As a result, current GRAG systems \cite{lightrag,edge2024localgraphrag} lack a structure-aware post-retrieval refinement mechanism with sufficient semantic modeling.
\textbf{(II) Structure-Reasoning Discrepancy.}
Incorporating structure embeddings (e.g., from GNNs) is hindered by a representation gap: graph embeddings  primarily capture structural signals, whereas LLMs operate over sequence semantics. This mismatch makes it difficult for LLM reasoning to reliably exploit graph structure, limiting graph-grounded generation \cite{liu2023multi,zhao2022learning}. Existing GRAG methods either linearize graphs as text \cite{tog,rog,xu2025harnessinglargelanguagemodels,GMT-KBQA,chen2024plan,Kg-agent} or incorporate graph encoders with simple concatenation/projection \cite{gnp,graphtoken,G-retriever}; however, they typically lack an explicit objective that aligns structural representations with the semantics required for LLM reasoning.
These research gaps present a question:

\definecolor{cvprpurple}{rgb}{0.52,0.34,0.74}
\begin{tcolorbox}[colframe=black!50, colback=cvprpurple!8, boxrule=1.5pt, arc=2mm, top=1pt, bottom=4pt, left=4pt, right=4pt, boxsep=1pt]
\raisebox{-0.2\baselineskip}{\includegraphics[height=1.1\baselineskip]{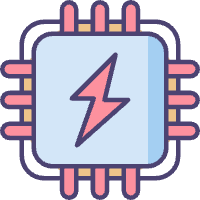}} 
\hspace{0.2em}
{\fontsize{10.8pt}{11.5pt}\selectfont\textit{How can we construct a compact yet structure-preserving subgraph and learn aligned graph representations for better LLM reasoning?}}
\end{tcolorbox}

In this work, we propose \ourmodel, a rationale-guided dual alignment framework tailored for GRAG. To tackle the two challenges above, we introduce a graph aligner as a pre-refinement module. Specifically, we leverage LLM with well-crafted prompts to extract rationale chains and anchors that highlight key intermediate concepts. We then use these signals as supervision to optimize dual alignment:  $\clubsuit\ $\textbf{Node-level Alignment}, which identifies and prioritizes reasoning-critical nodes/edges to prune structure-coupled irrelevant evidence; and $\spadesuit\ $\textbf{Graph-level Alignment}, which learns an aligned semantic space between graph and language representations via contrastive learning. Together, these components yield a compact yet structure-preserving subgraph, enabling the generator to produce more accurate and context-aware responses grounded in relational evidence.
Extensive experiments are conducted on GraphQA benchmark \cite{G-retriever}, covering commonsense reasoning, scene graph understanding, and knowledge graph reasoning tasks. The results consistently outperform 18 strong baselines and highlight the effectiveness of \ourmodel.

\begin{figure*}[t]
\centering
\includegraphics[width=0.9\linewidth]{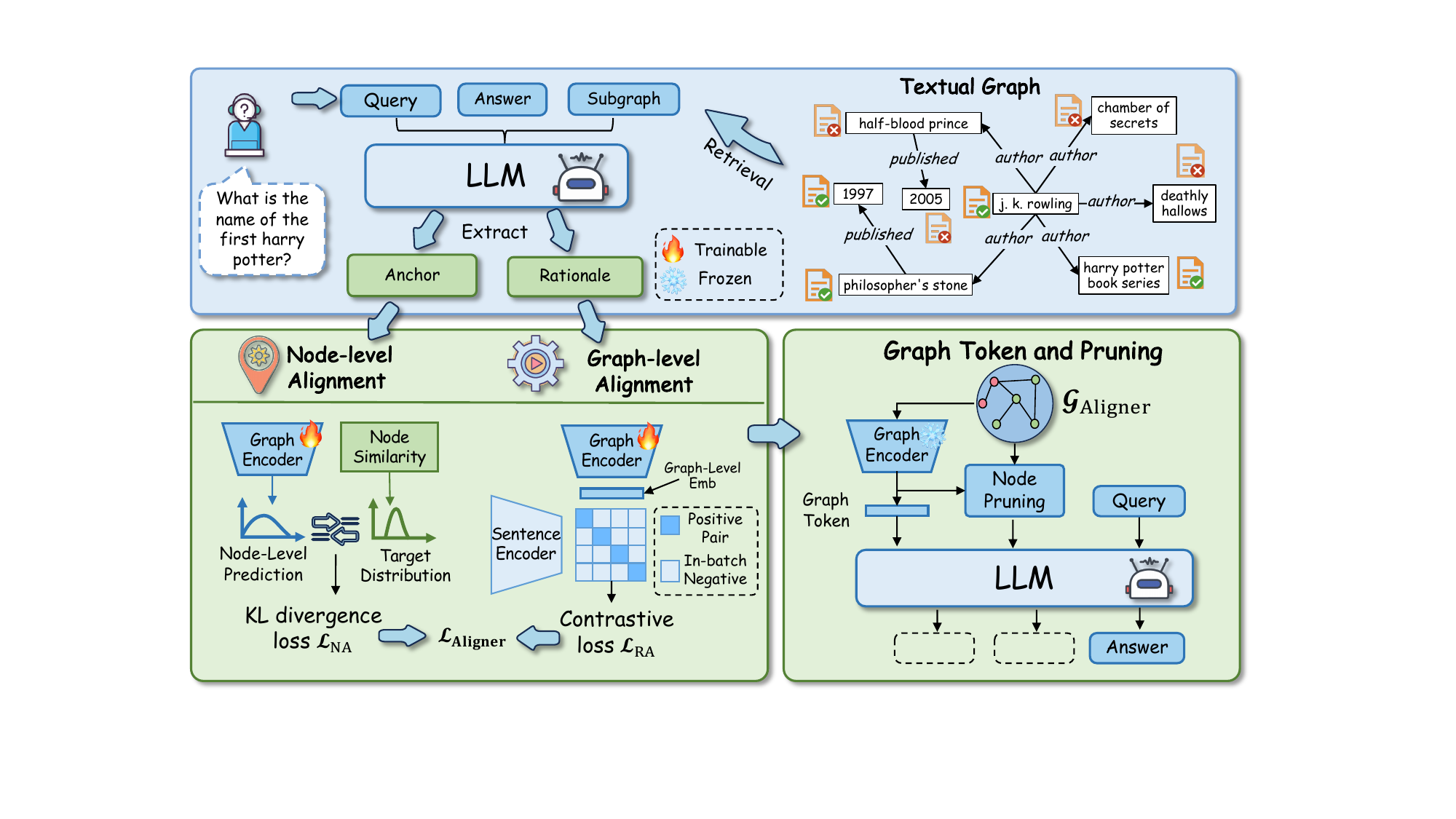}
\caption{Overview of \ourmodel. We first extract a rationale chain and anchors with LLM. The rationale guides graph-level alignment, while anchors supervise node-level alignment.}
\label{fig:method}
\end{figure*}

\section{Preliminaries}
\paragraph{Textual Graph.}
A textual graph is a graph where nodes and edges are enriched with textual information, capturing structural and semantic details. A textual graph is defined as $
\mathcal{G} = (\mathcal{V}, \mathcal{E}, \{t_n\}_{n \in \mathcal{V}}, \{t_e\}_{e \in \mathcal{E}})$,
where $ \mathcal{V} $ and $ \mathcal{E} $ are the sets of nodes and edges. $ t_n \in \mathcal{D}^{L_n} $ represents the text associated with a node $ n \in \mathcal{V} $, where $ \mathcal{D} $ is the vocabulary, and $ L_n $ is the length. Similarly, $ t_e \in \mathcal{D}^{L_e} $ is the text for edge $ e \in \mathcal{E} $ with length $ L_e $.

\paragraph{Task Formulation.}
This work addresses RAG with textual graph. The goal is to retrieve relevant information from textual graphs and generate accurate responses. Given a query $ t_q $, a sequence of tokens from $ \mathcal{D} $, the model retrieves a subgraph $ \mathcal{G}_r = (\mathcal{V}_r, \mathcal{E}_r) $ from $ \mathcal{G} $ based on the semantic similarity of $t_n$, $t_e$ with $t_q$, where $ \mathcal{V}_r \subseteq \mathcal{V} $ and $ \mathcal{E}_r \subseteq \mathcal{E} $. The retrieved nodes, edges, and texts are refined to improve the input quality for the LLM. During generation, the retrieved subgraph, query $ t_q $, and prompt $\mathcal{P}$ are provided as input to the LLM, which produces the final output $Y$, grounded in the retrieved knowledge and graph context.

\section{Methodology}
This section presents \ourmodel (Figure~\ref{fig:method}), an anchor-and-rationale guided refinement framework for GRAG. \ourmodel first extracts anchors concepts and compact rationale chains with LLM, and then uses them to guide a graph aligner that (i) performs structure-aware pruning to remove irrelevant evidence while preserving relational dependencies and (ii) reduces the gap between graph and language representations.

\subsection{\textbf{Anchor and Rationale Extraction}}\label{Extraction Generation}
To optimize the Aligner module, we leverage an explicit reasoning signal extracted for each query, which is especially important for multi-hop questions requiring intermediate steps. Inspired by Chain-of-Thought \cite{cot} and O1-style reasoning \cite{zhong2024evaluation}, we propose an LLM-based extraction module that produces (i) a compact rationale chain and (ii) a small set of anchors.

Specifically, given a question $q$ and its answer, we first retrieve an initial subgraph (as elaborated in Appendix~\ref{sec:Graph Retriever}). We then prompt a strong LLM to extract (i) a concise rationale chain that connects $q$ to the supporting evidence and (ii) anchors, i.e., key intermediate entities/relations that can be grounded to graph nodes, from the retrieved subgraph. For example, in Figure~\ref{fig:summarization_text}, for the query \textit{``What is the name of the first Harry Potter novel?''}, the extracted rationale highlights \texttt{J.K. Rowling} as a critical intermediate concept, which serves as an anchor. These outputs support dual alignment: the anchors \textbf{node-level alignment} for structure-aware pruning, while rationale guides supervise \textbf{graph-level alignment} to better ground language representations to graph.

\subsection{\textbf{Guided Dual Alignment}}
\paragraph{Node-level Alignment.} 
We propose an Aligner module that utilizes the extracted anchors to identify and align relevant nodes within the graph, effectively pruning redundant nodes and edges by filtering out unrelated information at the node level.
 Specifically, we employ a GNN (e.g., GraphTransformer \cite{graphtransformer} or GAT \cite{gat}), to encode the structural information of the graph. The GNN produces node-level embeddings $\boldsymbol{n}_g$ based on the input graph:
\begin{equation}
    \boldsymbol{n}_g = \text{GNN}(\mathcal{G}) \in \mathbb{R}^{|\mathcal{V}|\times d},
\end{equation}
where $|\mathcal{V}|$ is the number of node and $d$ is feature dim.
For the anchors, we employ SBERT to encode the textual description into a embedding $\boldsymbol{r}_s$ that captures its semantic meaning:
\begin{equation}
    \boldsymbol{r}_s = \text{SBERT}(t_{\text{anchor}}) \in \mathbb{R}^{d_s}.
\end{equation}
 We concatenate the embedding of each node in the subgraph with the query embedding. The concatenated embeddings are then passed through an MLP module, which generates a predicted importance score for each node $\boldsymbol{p}_\text{prediction}$. These scores are transformed into probability distributions using the softmax function, which produces importance scores for both the prediction and the anchor. 
\begin{align}
    \boldsymbol{p}_\text{prediction} &= \text{Softmax}(\text{MLP}([\boldsymbol{n}_g, \boldsymbol{q}^\text{expand}]))  \in \mathbb{R}^{|\mathcal{V}|}, \\
    \boldsymbol{p}_\text{anchor} &= \text{Softmax}(\text{cos}(\boldsymbol{n}_t,\boldsymbol{r}_s)) \in \mathbb{R}^{|\mathcal{V}|},
\end{align}
where $[, ]$ means concat operation, $\text{cos()}$ means cosine similarity, $\boldsymbol{n}_t$ is text embedding of node, $\boldsymbol{q}^\text{expand}$ is the query embedding broadcasted across all nodes.
To align the node distribution, we minimize the Kullback–Leibler (KL) divergence \cite{kullback1951information} between predicted probabilities $\boldsymbol{p}_\text{prediction}$ and the probabilities $\boldsymbol{p}_\text{anchor}$. The KL divergence loss for a subgraph is given by:
\begin{equation}
    \mathcal{L}_\text{NA} =\frac{1}{|\mathcal{V}|} \sum_{i=1}^{|\mathcal{V}|} \boldsymbol{p}_\text{anchor}(i) \log \frac{\boldsymbol{p}_\text{anchor}(i)}{\boldsymbol{p}_\text{prediction}(i)}.
\end{equation}
Optimizing $\mathcal{L}_\text{NA}$ enables effective alignment of relevant knowledge.

\paragraph{Graph-level Alignment.} 
To bridge the representation gap between graph structures and textual descriptions, our Aligner module treats the text representation derived from rationale chains as the target label, aligning the graph and text embeddings to encourage semantic consistency.
We apply a mean pooling operation across the node embeddings $\boldsymbol{n}_g$ to obtain a unified graph-level token $\boldsymbol{r}_g$:
\begin{equation}
\boldsymbol{r}_g = \text{POOL}(\boldsymbol{n}_g) = \frac{1}{|\mathcal{V}|} \sum_{v \in \mathcal{V}} \boldsymbol{n}_g(v) \in \mathbb{R}^{d}.
\end{equation}
To unify the graph and text representations in a shared semantic space, we apply a contrastive loss with in-batch negative sampling. This loss encourages positive pairs (i.e., graph and text embeddings) to have higher similarity, while pushing apart non-matching pairs. Then, a shared-weight MLP layer is further designed to map $\boldsymbol{r}_g$ and $\boldsymbol{r}_s$ to dimension $d_t$ of LLM token embeddings: 
\begin{equation}
\boldsymbol{\hat{r}}_s = \text{MLP}(\boldsymbol{r}_s), \quad \boldsymbol{\hat{r}}_g = \text{MLP}(\boldsymbol{r}_g) .
\end{equation}
The contrastive loss for graph-level alignment from $\boldsymbol{\hat{r}}_g $ to $\boldsymbol{\hat{r}}_s$ is defined as:
\begin{equation}
\small
    \mathcal{L}_{GA(\boldsymbol{\hat{r}}_g \rightarrow \boldsymbol{\hat{r}}_s)} = -\frac{1}{N} \sum_{i=1}^N 
    \left[ 
    \log \frac{\exp\left( \text{cos}(\boldsymbol{\hat{r}}_g^i, \boldsymbol{\hat{r}}_s^i) / \tau \right)}{\sum_{j=1}^N \exp\left( \text{cos}(\boldsymbol{\hat{r}}_g^i, \boldsymbol{\hat{r}}_s^j) / \tau \right)}
    \right],
\end{equation}
where, $N$ is the batch size, $(\boldsymbol{\hat{r}}_g^i, \boldsymbol{\hat{r}}_s^i)$ is the $i$-th positive (graph-text) pair in the batch, $\tau$ is a temperature parameter to control the sharpness. Similarly, we can obtains the loss $\mathcal{L}_{GA( \boldsymbol{\hat{r}}_s \rightarrow \boldsymbol{\hat{r}}_g )}$ from $\boldsymbol{\hat{r}}_s$ to $\boldsymbol{\hat{r}}_g $. The final representation alignment loss is obtained by:
\begin{equation}
\mathcal{L}_{GA} = \frac{1}{2}\Big(
\mathcal{L}_{GA(\boldsymbol{\hat{r}}_s \rightarrow \boldsymbol{\hat{r}}_g)}
+
\mathcal{L}_{GA(\boldsymbol{\hat{r}}_g \rightarrow \boldsymbol{\hat{r}}_s)}
\Big).
\end{equation}
To achieve an optimized Graph Aligner, we perform joint optimization of node and graph Alignment. The total loss for the Graph Aligner is defined as:
$ \mathcal{L}_\text{Aligner} = \mathcal{L}_\text{GA} + \mathcal{L}_\text{NA}
$. The parameters of GNN encoder are jointly optimized using the loss $\mathcal{L}_\text{Aligner}$ with a specified training step, which reflects degree of alignment. We evaluate the impact of the alignment degree in Figure \ref{fig:align} and the effectiveness in Section \ref{sec:Representation Alignment}.

\subsubsection{\textbf{Graph Token and Pruning}} \label{Graph Pruning and Representation Generation}

We apply the trained Graph Aligner to obtain an aligned graph representation token and a pruning subgraph for generation. Note that gold answers are only used to train the Aligner.

\paragraph{Encoding graph token.}
Given the retrieved subgraph, we leverage the Aligner to produce node embeddings and a pooled graph token $\boldsymbol{r}_g$, which is optimized by $\mathcal{L}_\text{Aligner}$ to be consistent with the rationale-derived text space.

\paragraph{Structure-aware pruning.}
We then use the node-level prediction scores to remove irrelevant evidence. Specifically, we select the top $n_{\text{seed}}$ nodes as seed nodes and expand them with their first-order neighbors to form the pruned subgraph $\mathcal{G}_\text{Aligner} = (\mathcal{V}_\text{Aligner}, \mathcal{E}_\text{Aligner})$.

Finally, we concatenate the linearized text of $\mathcal{G}_\text{Aligner}$ with the query tokens $t_q$ and feed them into the LLM embedding layer to obtain token embeddings $\boldsymbol{r}_t$. We fuse $\boldsymbol{r}_t$ with $\boldsymbol{r}_g$ to condition answer generation:
\begin{equation}
\small
p_\theta(Y \mid t_q, \mathcal{G}_\text{Aligner})=\prod_{i=1}^{m} p_\theta\!\left(y_i \mid [\boldsymbol{r}_g,\boldsymbol{r}_t],\, y_{<i}\right).
\end{equation}
For efficiency, we fine-tune the generator with parameter-efficient methods such as LoRA~\cite{lora}.

% SUMMARIZE\_PROMPT = f"""
% You are a helpful assistant responsible for generating a comprehensive summary of the data provided below.
% Given question and answer, and related graph data base.
% Please concatenate all of these into a single, comprehensive description. The description should logically connect the question to the answer. Make sure to include information collected from all descriptions.
% If the provided descriptions are contradictory, please resolve the contradictions and provide a single, coherent summary.
% Make sure it is written in third person, and include the entity names so we have the full context.

% \#\#\#\#\#\#\#
% -Data-
% Question: \{data['question']\}
% Answer: \{truncate\_string(data['full\_answer'],128)\}
% Graph Data Base: \{truncate\_string(data['cached\_desc'],2048)\}
% \#\#\#\#\#\#\#
% Output:
% """

% \section{Experiments}
% We conducted experiments on the GraphQA benchmark \cite{G-retriever}, which includes ExplaGraphs (commonsense reasoning), SceneGraphs (scene graph understanding), and WebQSP (knowledge graph reasoning). To ensure a fair comparison, we utilized the same retrieval results obtained via the PCST algorithm across all baselines. More details about datasets, baselines and implementation details are provided in Appendix \ref{sec:Experimental Settings}.

\lstdefinestyle{prompt}{
    basicstyle=\ttfamily\fontsize{7pt}{8pt}\selectfont,
    frame=none,
    breaklines=true,
    backgroundcolor=\color{lightgray},
    breakatwhitespace=true,
    breakindent=0pt,
    escapeinside={(*@}{@*)},
    numbers=none,
    numbersep=5pt,
    xleftmargin=5pt,
    aboveskip=2pt,
    belowskip=2pt,
}
\tcbset{
  aibox/.style={
    top=10pt,
    colback=white,
    % colframe=black,
    % colbacktitle=black,
    enhanced,
    center,
    % attach boxed title to top left={yshift=-0.1in,xshift=0.15in},
    % boxed title style={boxrule=0pt,colframe=white,},
  }
}
\newtcolorbox{AIbox}[2][]{aibox, title=#2,#1}

\section{Experiments}
\subsection{Experimental Settings} \label{sec:Experimental Settings}

\begin{table*}[t]
\small
\setlength\tabcolsep{7.pt}  %可以控制列间距
\centering
\begin{tabular}{l|ccccc}
\toprule
 \multicolumn{1}{c|}{\multirow{2}{*} {\textbf{Method}}} & \textbf{ExplaGraphs}  & \textbf{SceneGraphs} & \multicolumn{3}{c}{\textbf{WebQSP}} \\
\cline{2-6}
  \multicolumn{1}{c|}{} & Accuracy$\uparrow$ & Accuracy$\uparrow$ &F1$\uparrow$&Hit@1$\uparrow$& Accuracy$\uparrow$ \\ 
\midrule
\multicolumn{6}{c}{\textbf{\textit{\cellcolor{gray!0}Inference-only}}} \\
\midrule
 Zero-shot$\ddagger$ & 0.5650 &0.3974&-&0.4106&-\\
 Zero-CoT$\ddagger$ & 0.5704 &0.5260 &-&0.5130&-\\
 CoT-BAG$\ddagger$ & 0.5794 &0.5680&-&0.3960&-\\
 KAPING$\ddagger$ & 0.6227 &0.4375 &-&0.5264&-\\
ToG & 0.7664$\pm$ 0.0348 & 0.6194$\pm$0.0460 & 0.3014$\pm$ 0.0245 & 0.5871$\pm$ 0.0258 & 0.3719$\pm$0.0183\\
\midrule
\multicolumn{6}{c}{\textbf{\textit{\cellcolor{gray!0}Raw Fine-tuning}}} \\
\midrule
 Prompt tuning& 0.5763$\pm$0.0243 &0.6341$\pm$0.0024 & 0.2652$\pm$0.0049 & 0.4807$\pm$0.0055 & 0.2827$\pm$0.0073\\
 LoRA & 0.8538$\pm$0.0353 & 0.7862$\pm$0.0031 & 0.4445$\pm$0.0058 & 0.6505$\pm$0.0068 & 0.4479$\pm$0.0091\\
\midrule
\multicolumn{6}{c}{\textbf{\textit{\cellcolor{gray!0}Reranker-based}}} \\
\midrule
 GTE-base & 0.8557$\pm$0.0144& 0.8556$\pm$0.0095 & 0.5378$\pm$0.0044 & 0.7373$\pm$0.0064 & 0.5251$\pm$0.0052 \\
 GTE-large & 0.8776$\pm$0.0095&  0.8592$\pm$0.0074 &\underline{0.5392$\pm$0.0013} & 0.7340$\pm$0.0044 & 0.5374$\pm$0.0038\\
 BGE-reranker-base & 0.8534$\pm$0.0159&  0.8577$\pm$0.0029 & 0.5323$\pm$0.0052 & 0.7397$\pm$0.0012 & 0.5254$\pm$0.0010 \\
 BGE-reranker-large & 0.8612$\pm$0.0184&  0.8644$\pm$0.0060 & 0.5366$\pm$0.0045 & 0.7391$\pm$0.0093 & 0.5401$\pm$0.0077 \\
 G-RAG &  0.8484$\pm$0.0174 &  0.8474$\pm$0.0147 & 0.5181$\pm$0.0023 & 0.7114$\pm$0.0113 & 0.5080$\pm$0.0041\\
 G-RAG-RL &  0.8478$\pm$0.0112 &  0.8509$\pm$0.0142 & 0.5291$\pm$0.0066 & 0.7167$\pm$0.0039 & 0.5185$\pm$0.0026\\
\midrule
\multicolumn{6}{c}{\textbf{\textit{\cellcolor{gray!0}GNN-based}}} \\
\midrule
 GraphToken$\ddagger$  & 0.8508$\pm$ 0.0551 &0.4903$\pm$ 0.0105 &-&0.5705$\pm$0.0074&-\\
 GNP  & 0.8704$\pm$0.0034&  0.8616$\pm$0.0096 &0.5369$\pm$0.0049 & 0.7391$\pm$0.0100 & 0.5441$\pm$0.0046 \\
 $\text{G-Retriever}_{PT}^\dagger$ & 0.8516$\pm$0.0092 & 0.8131$\pm$0.0162 & 0.4740$\pm$0.0049 & 0.6921$\pm$0.0099 & 0.4740$\pm$0.0033\\
 $\text{G-Retriever}_{LoRA}^\dagger$& 0.8705$\pm$0.0329 & \underline{0.8683$\pm$0.0072}& 0.5366$\pm$0.0031 & 0.7366$\pm$0.0049 & 0.5405$\pm$0.0031\\
 $\text{GRAG}_{PCST}^\dagger$   &\underline{0.8805$\pm$0.0050}& 0.8561$\pm$0.0052&0.5355$\pm$0.0049&\underline{0.7485$\pm$0.0104}&\underline{0.5503$\pm$0.0035}\\
 \cellcolor{blue!10}\textbf{\ourmodel (Ours)} & \cellcolor{blue!10}\textbf{0.8992$\pm$0.0124} &  \cellcolor{blue!10}\textbf{0.8804$\pm$0.0106} & \cellcolor{blue!10}\textbf{0.5445$\pm$0.0041} & \cellcolor{blue!10}\textbf{0.7626$\pm$0.0063} & \cellcolor{blue!10}\textbf{0.5700$\pm$0.0039}\\
 \cellcolor{blue!10}\textbf{Improvement $(\triangle)$}&\cellcolor{blue!10}  +2.12\%{\Large *} &\cellcolor{blue!10} +1.39\% &\cellcolor{blue!10}+1.68\%  &\cellcolor{blue!10}+1.88\%{\Large *}&\cellcolor{blue!10}+3.58\%{\Large *}\\
\bottomrule
\end{tabular}
\caption{Performance comparison using the same retrieval settings and generator for all methods. The table reports the mean and standard deviation results across three random seeds. For methods marked with `$\dagger$', we reproduce the results on WebQSP to report F1 and Accuracy. Results for methods marked with `$\ddagger$' are taken directly from \citet{G-retriever}. The best results are highlighted in \textbf{bold}, and the second-best results are \underline{underlined}. `Improvement $(\triangle)$' represents the gain over second-best baseline.   `\textbf{{\Large *}}' indicates statistically significant improvements (i.e., two-sided t-test with $p<0.05$) over second-best baseline. $\uparrow$: higher is better.}
\label{tab:main result}
\end{table*}

\paragraph{Datasets.}
Following G-Retriever \cite{G-retriever}, we conducted experiments on the GraphQA benchmark \cite{G-retriever}, which includes ExplaGraphs (commonsense reasoning), SceneGraphs (scene graph understanding), and WebQSP (knowledge graph reasoning). The dataset statistics are summarized in Appendix~\ref{sec:appendix_dataset}.

\paragraph{Baselines.}
We group baselines into four categories. (1) \textbf{Inference-only} methods leverage frozen LLMs by taking textualized graph and query as input, including Zero-shot, Zero-CoT~\cite{cot}, CoT-BAG~\cite{wang2024can}, KAPING~\cite{baek2023knowledge}, and ToG~\cite{tog}. (2) \textbf{Raw Fine-tuning} adapts LLM via parameter-efficient tuning, covering Prompt Tuning~\cite{pt} and LoRA~\cite{lora}. (3) \textbf{Reranker-based} baselines rerank retrieved documents before feeding them into LLM, including GTE~\cite{gte}, BGE~\cite{bge}, G-RAG, and G-RAG-RL~\cite{dong2024don}. (4) \textbf{GNN-based} approaches integrate GNN encoders with LLM embeddings, including GraphToken~\cite{graphtoken}, GNP~\cite{gnp}, G-Retriever~\cite{G-retriever}, and GRAG~\cite{hu2024grag}.

\paragraph{Implementation Details.} \label{sec:implementation_details}
We use GraphTransformer \cite{graphtransformer}, GAT \cite{gat}, and GCN \cite{gcn} as GNN encoders, and Llama-2-7b-hf, Llama-2-13b-hf and Llama-3.1-8b \cite{touvron2023llama} as generators. For the retrieval process, we use the same retrieval results across all baselines to ensure fair comparisons. Both reranker-based and GNN-based methods apply LoRA for fine-tuning. All methods are compared using the same training hyperparameters where applicable. In our module, we explore two hyperparameters: alignment degree and the number of seed nodes ($n_\text{seed}$), with the analysis shown in Figure \ref{fig:hyperpara_topk}.
To extract the anchor and rationale, we employ Llama-3.1-70B-Instruct \cite{dubey2024llama}.
More details can be found in Appendix \ref{app:experimentaldetails}.

\begin{table*}[!ht]
\setlength\tabcolsep{3pt}
\small
\centering
\begin{tabular}{c|l|cc|cc|cc|cc}
\toprule
\multicolumn{2}{c|}{\textbf{LLM Backbones}} &  \multicolumn{2}{c|}{\textbf{Llama-2-7b-hf}} &\multicolumn{2}{c}{\textbf{Llama-2-13b-hf}} &\multicolumn{2}{c}{\textbf{Llama-3.1-8B}}  &\multicolumn{2}{c}{\textbf{Qwen3-8B}} \\
\midrule
\multicolumn{1}{c|}{\multirow{2}{*} {\textbf{GNN} }} & \multicolumn{1}{c|}{\multirow{2}{*} {\textbf{Method}}} & \textit{ExplaGraphs}  & \textit{WebQSP} & \textit{ExplaGraphs}  & \textit{WebQSP} & \textit{ExplaGraphs}  & \textit{WebQSP} & \textit{ExplaGraphs}  & \textit{WebQSP}\\
 \multicolumn{1}{c|}{} & \multicolumn{1}{c|}{} &Accuracy$\uparrow$ &Hit@1$\uparrow$ &Accuracy$\uparrow$ &Hit@1$\uparrow$&Accuracy$\uparrow$ &Hit@1$\uparrow$ &Accuracy$\uparrow$ &Hit@1$\uparrow$\\ 
\midrule

\multicolumn{1}{c|}{\multirow{3}{*} {GT}} & GNP & 0.8704 & 0.7391 &0.8880 & 0.7696 & 0.8892 & 0.7557 & 0.8915 & 0.7602 \\
\multicolumn{1}{c|}{}& G-Retriever& 0.8705 & 0.7366& 0.9115 & 0.7739 & 0.9014 & 0.7482 & 0.9088 & 0.7535 \\
\multicolumn{1}{c|}{}& \cellcolor{blue!10}\textbf{\ourmodel} & \cellcolor{blue!10}\textbf{0.8992}& \cellcolor{blue!10}\textbf{0.7626}& \cellcolor{blue!10}\textbf{0.9241} & \cellcolor{blue!10}\textbf{0.7789} & \cellcolor{blue!10}\textbf{0.9321} & \cellcolor{blue!10}\textbf{0.7712} & \cellcolor{blue!10}\textbf{0.9365} & \cellcolor{blue!10}\textbf{0.7794} \\
\midrule
\multicolumn{1}{c|}{\multirow{3}{*} {GAT}} & GNP &  0.9061 & 0.7291& 0.8989 & 0.7676 & 0.9062 & 0.7521 & 0.9104 & 0.7588 \\
\multicolumn{1}{c|}{}& G-Retriever&0.7960 & \textbf{0.7414}& 0.8953 & \textbf{0.7737} & 0.9111 & 0.7719 & 0.9150 & 0.7760 \\
\multicolumn{1}{c|}{}& \cellcolor{blue!10}\textbf{\ourmodel} & \cellcolor{blue!10}\textbf{0.9151}& \cellcolor{blue!10}0.7309 & \cellcolor{blue!10}\textbf{0.9151} & \cellcolor{blue!10}0.7573 & \cellcolor{blue!10}\textbf{0.9214} & \cellcolor{blue!10}\textbf{0.7742} & \cellcolor{blue!10}\textbf{0.9282} & \cellcolor{blue!10}\textbf{0.7815} \\
\midrule
\multicolumn{1}{c|}{\multirow{3}{*} {GCN}} & GNP &0.7545 & 0.7298& 0.8682 & 0.7564 & 0.8572 & 0.7495 & 0.8615 & 0.7530 \\
\multicolumn{1}{c|}{}& G-Retriever& \textbf{0.8592}& 0.7352&0.9007 & 0.7521 & 0.8512& 0.7527 & 0.8588 & 0.7592 \\
\multicolumn{1}{c|}{}& \cellcolor{blue!10}\textbf{\ourmodel} & \cellcolor{blue!10}0.8574&\cellcolor{blue!10}\textbf{0.7377}& \cellcolor{blue!10}\textbf{0.9152}& \cellcolor{blue!10}\textbf{0.7592} & \cellcolor{blue!10}\textbf{0.8596} & \cellcolor{blue!10}\textbf{0.7573} & \cellcolor{blue!10}\textbf{0.8695} & \cellcolor{blue!10}\textbf{0.7645} \\
\bottomrule
\end{tabular}
\caption{Generalization Analysis on different LLM and GNN backbones.}
\label{tab:Generalization}
\end{table*}

\subsection{Main Results}
We conduct extensive experiments against 18 baselines, and Table \ref{tab:main result} shows that \ourmodel consistently improves all metrics across three datasets, with a particularly large Accuracy gain on WebQSP over the second-best method due to its alignment-based graph pruning and optimization. Compared with \textbf{inference-only methods}, which rely purely on LLM reasoning (e.g., Zero-shot) or heuristic KG search (e.g., ToG), \ourmodel benefits from task-aware structural optimization and achieves clearly stronger results. Compared with \textbf{raw fine-tuning methods} such as Prompt Tuning and LoRA, \ourmodel further demonstrates that parameter-efficient adaptation alone is insufficient to fully exploit graph structure. Compared with \textbf{reranker-based methods} (e.g., gte-large and related variants), which mainly rerank nodes/triples and thus miss richer structural dependencies, \ourmodel achieves superior performance by aligning graph reasoning with LLM-derived reasoning chains as the optimization target. Compared with \textbf{GNN-based methods} (e.g., G-Retriever and GNP), which inject graph embeddings via projection or pooling but lack explicit graph--language alignment, \ourmodel more effectively bridges the two representation spaces and yields state-of-the-art performance.

\subsection{Generalization Analysis}
In this section, we evaluate \textbf{whether \ourmodel generalizes across both GNN architectures and LLM scales} on ExplaGraphs (Accuracy) and WebQSP (Hit@1). As summarized in Table \ref{tab:Generalization}, across three GNN backbones (GT, GAT, GCN) and multiple LLM families/sizes, \ourmodel outperforms GNP and G-Retriever in the vast majority of configurations, indicating strong robustness to different encoder--reasoner pairings. The largest gains are observed with the GraphTransformer backbone, suggesting that its ability to capture long-range dependencies synergizes with \ourmodel’s alignment objectives to improve retrieval and reasoning, while consistent improvements with GAT further show that alignment complements attention-based aggregation. Moreover, scaling up the LLM generally boosts results—most notably with the weaker GCN encoder—implying that larger LLMs can better leverage aligned graph signals to compensate for limited graph expressivity. The consistent gains on both datasets confirm that \ourmodel transfers well across different settings.

\subsection{Ablation Study}

% \begin{figure}[htp]
% \centering
% \includegraphics[width=0.2\columnwidth]{img/ablation-Accuracy.pdf}
% \includegraphics[width=0.2\columnwidth]{img/ablation-Hit-1.pdf}
% \includegraphics[width=0.2\columnwidth]{img/ablation-F1.pdf}
% \includegraphics[width=0.3\columnwidth]{img/ablation-legend.pdf}
% \caption{Ablation study with different alignment strategy.}
% \label{fig:ablation}
% \end{figure}

We conduct an ablation study on WebQSP with three metrics to assess the contribution of each module. As shown in Table~\ref{tab:ablation}, we consider four variants: \textbf{(1) w/o Node Alignment} (removing the KL loss and pruning), \textbf{(2) w/o Graph Alignment} (removing the contrastive loss and graph token), \textbf{(3) w/o Both}, and \textbf{(4) Random Alignment} (training the aligner with random labels). Removing Node Alignment causes the largest degradation across all metrics, confirming the importance of node-level alignment and pruning. Disabling Graph Alignment also reduces performance, and removing both modules further amplifies the drop, underscoring the necessity of dual alignment for suppressing irrelevant knowledge and mitigating the graph language representation gap. Finally, Random Alignment performs worst, indicating that effective alignment requires meaningful supervision; our LLM-extracted anchor and rationale provide such a useful signal.

\begin{table}[t]
\small
\centering
\begin{tabular}{l|ccc}
\toprule
 \multicolumn{1}{c|}{\multirow{2}{*} {\textbf{Method}}} & \multicolumn{3}{c}{\textbf{WebQSP}} \\
\cline{2-4}
  \multicolumn{1}{c|}{} &F1$\uparrow$&Hit@1$\uparrow$& Accuracy$\uparrow$ \\ 
\midrule
  \cellcolor{blue!10}\textbf{\ourmodel} & \cellcolor{blue!10}\textbf{0.5445} & \cellcolor{blue!10}\textbf{0.7626} &  \cellcolor{blue!10}\textbf{0.5700}\\
w/o Graph Alignment &0.5423 &0.7578 &0.5673 \\
w/o Node Alignment &0.5344 &0.7371 &0.5339 \\
w/o Both &0.5348 &0.7328 &0.5216\\
Random Alignment &0.4617 &0.6861 &0.4865\\
\bottomrule
\end{tabular}
\caption{Ablation study on different alignment strategy.}
\label{tab:ablation}
\end{table}

\begin{figure*}[t]
\centering
\begin{subfigure}[t]{0.68\columnwidth}
    \centering
    \includegraphics[width=\textwidth]{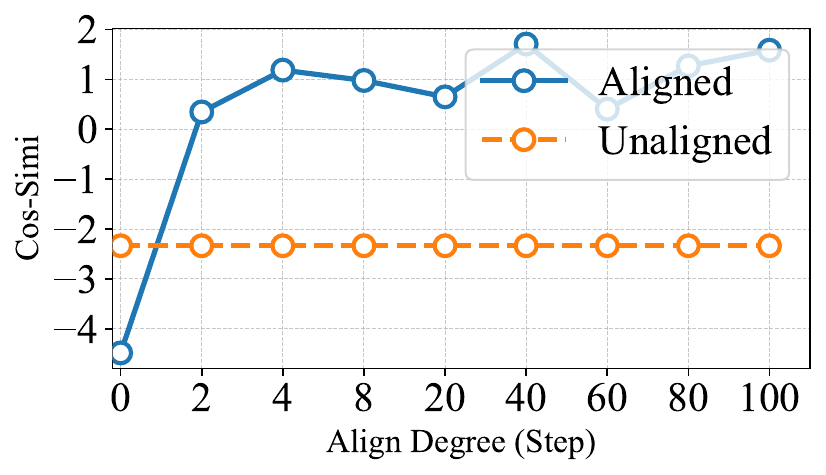}
    \caption{Align with Query}
\end{subfigure}%
\hfill
\begin{subfigure}[t]{0.68\columnwidth}
    \centering
    \includegraphics[width=\textwidth]{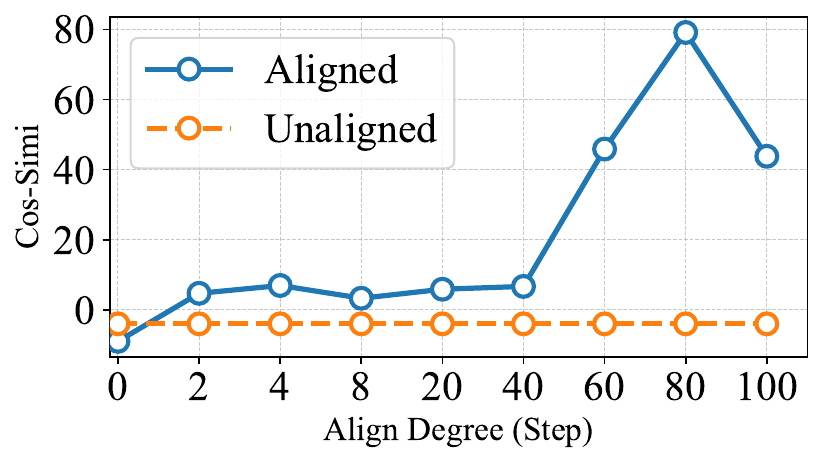}
    \caption{Align with Rationale}
\end{subfigure}%
\hfill
\begin{subfigure}[t]{0.68\columnwidth}
    \centering
    \includegraphics[width=\textwidth]{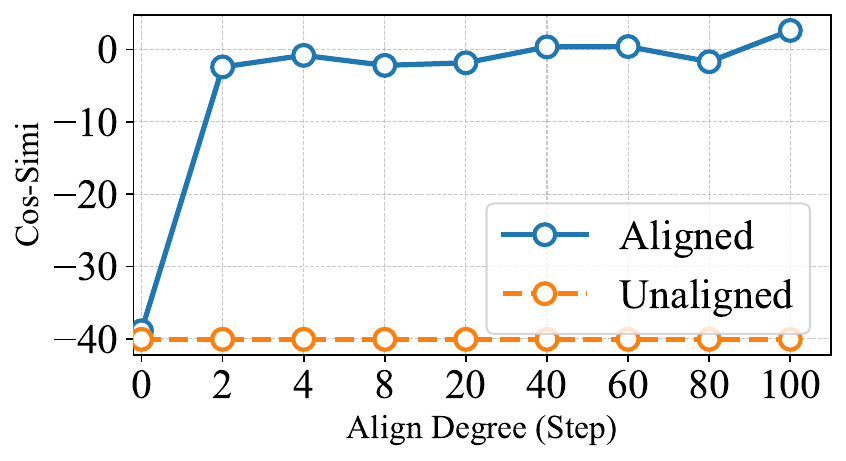}
    \caption{Align with Textualized Graph}
\end{subfigure}
\caption{Graph-level Alignment Analysis: The cosine similarity score between graph embeddings and language embeddings (aligned using the aligner module vs. the unaligned setting).}
\label{fig:align}
\end{figure*}

\subsection{Evaluation of Graph-level Alignment} \label{sec:Representation Alignment}

In this section, we evaluate \textbf{whether the aligner can effectively bridge the representation gap?} Specifically, we measure cosine similarity between graph embeddings (unaligned vs.\ aligned via our contrastive loss) and language embeddings of the query, rationale, and textualized graph on the test set; Figure \ref{fig:align} shows a clear upward trend in similarity as alignment proceeds, with aligned embeddings consistently surpassing unaligned ones after sufficient training, indicating that the aligner effectively narrows the graph--language representation gap. Meanwhile, as discussed in Section \ref{sec:Impact of Align Degree}, the benefit saturates beyond a certain point, since overly strong alignment can distort original graph information and hurt accuracy.

\begin{figure}[t]
\centering
\includegraphics[width=0.84\linewidth]{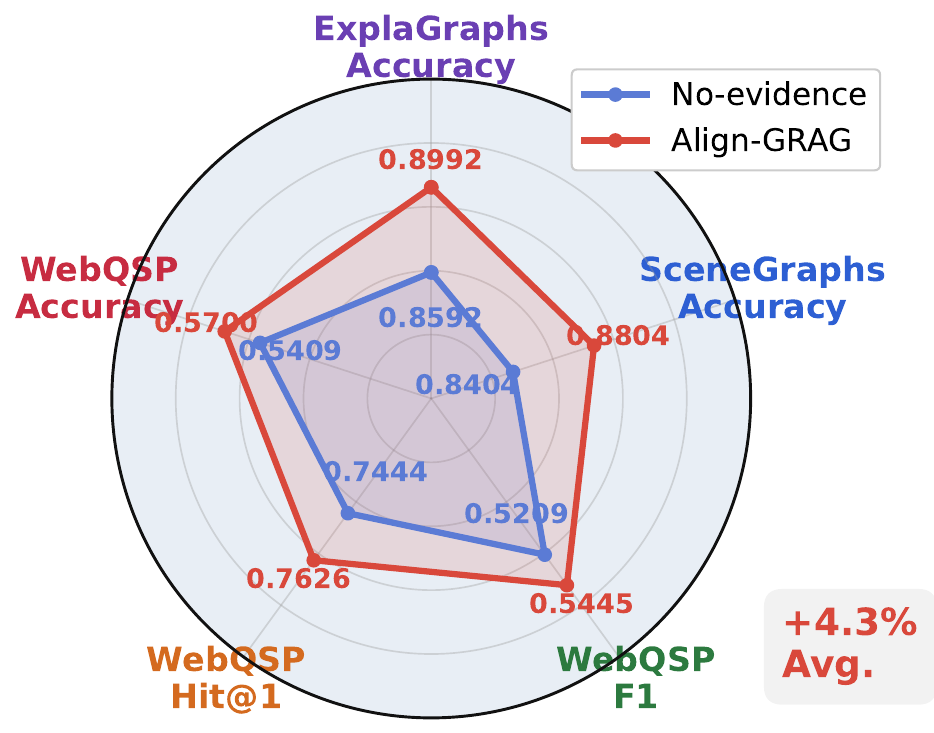}
\caption{Quantitative Comparison of Different anchor and rationale Quality.}
\label{fig:quantitative comparison}
\end{figure}

\subsection{\textbf{Quality Analysis}}
We investigate \textbf{whether the quality of extracted anchors and rationale directly affects QA performance.} We compare \ourmodel with \texttt{No-evidence Extraction}, where the LLM generates anchors and a rationale chain using only the question (without accessing the retrieved subgraph evidence). As shown in Figure~\ref{fig:quantitative comparison}, \ourmodel consistently performs better across datasets and metrics. This gap indicates that evidence-grounded extraction produces more relevant and faithful supervision signals for both node- and graph-level alignment. In contrast, when extraction is detached from graph evidence, the generated chains are more prone to plausible-but-ungrounded reasoning, which weakens pruning and representation alignment, ultimately harming QA.

\colorlet{pruneA}{orange!5}
\colorlet{pruneB}{orange!12}
\colorlet{pruneC}{red!12}
\colorlet{pruneD}{red!25}
\colorlet{pruneE}{red!30}
\begin{table}[t]
\setlength\tabcolsep{2pt}  %可以控制列间距
\small
\centering
\begin{tabular}{l|ccc}
\toprule
\textbf{Method} & \# \textbf{Tokens} & \textbf{Infer Time} & \textbf{Hit@1} \\
\midrule
\multicolumn{4}{c}{\textbf{No Pruning}} \\
\midrule
Non-Retriever & 100626.28 (387.02\%) & OOM & - \\
BM25         & 2569.57 (99.72\%)    & 17:24 min & 0.4394 \\
Contriever   & 2593.28 (100.01\%)   & 17:37 min & 0.4429 \\
G-Retriever  & 2576.66 (100.00\%)   & 17:28 min & 0.4502 \\
\midrule
\multicolumn{4}{c}{\textbf{Node Pruning with Different Degree}} \\
\midrule
\multicolumn{4}{l}{\textbf{\ourmodel}}  \\
\rowcolor{pruneA} \quad $n_\text{seed}=4$  & 496.54 (19.27\%)  & 3:31 min  & 0.4299$\downarrow$ \\
\rowcolor{pruneB} \quad  $n_\text{seed}=6$  & 698.54 (27.11\%)  & 4:45 min  & \textbf{0.4527$\uparrow$} \\
\rowcolor{pruneC} \quad  $n_\text{seed}=8$  & 905.52 (35.14\%)  & 6:55 min  & \textbf{0.4699$\uparrow$} \\
\rowcolor{pruneD} \quad $n_\text{seed}=10$ & 1120.75 (43.50\%) & 8:27 min  & \textbf{0.4785$\uparrow$} \\
\rowcolor{pruneE} \quad $n_\text{seed}=15$ & 1546.90 (60.04\%) & 11:50 min & \textbf{0.4914$\uparrow$} \\
\bottomrule
\end{tabular}
\caption{Evaluation of Efficiency with and without node pruning for inference time. $\uparrow$ indicates results better than G-Retriever, while $\downarrow$ indicates worse results. (xx\%) represents the percentage of tokens relative to G-Retriever. OOM means out of memory.}
\label{tab:inferencetime}
\end{table}

\subsection{Evaluating the Efficiency of Node Pruning}\label{sec:Evaluation of Efficiency}

In this section, we evaluate \textbf{the efficiency improvements brought by node pruning in Aligner module}. In an inference-only setting on WebQSP, we compare retrieval variants (BM25, Contriever, G-Retriever, and w/ Aligner with different seed-node choices). To fully utilize resources and memory, we adjusted the batch size to be as large as possible under different average token conditions. For comparison, the BM25 and Contriever methods retrieve triples and ensure that the token consumption is roughly similar to that of G-Retriever. Table \ref{tab:inferencetime} shows that with a moderate number of seed nodes, our pruning preserves core evidence while removing redundant context—\textbf{achieves performance comparable to G-Retriever while only utilizing 27.11\% tokens}, whereas too few seeds can over-prune and hurt performance, and more seeds retain more useful knowledge to outperform G-Retriever in longer-context regimes, indicating that selective pruning improves both efficiency and effectiveness. Finally, under similar token budgets, BM25 and Contriever lag behind, suggesting that graph-aware retrieval better exploits relational connectivity.

% \subsection{\textbf{Evaluation of Relevance and Faithfulness}}
% We evaluate the extracted anchors and rationale chains for (i) relevance to the question answer context and (ii) faithfulness to the provided evidence, following an LLM-as-a-judge protocol \cite{gu2024survey}. We employ GPT-4o, Gemini-2.5-Flash, and human annotators, and randomly sample 100 instances. The evaluation prompt is shown in Figure~\ref{fig: Relevance and Faithfulness Prompt}, and the results are reported in Figure~\ref{fig:Evaluation of Relevance and Faithfulness}.
% Overall, the summaries demonstrate strong quality. Relevance reaches 90\%/92\%/92\% and faithfulness reaches 87\%/89\%/90\% under GPT-4o, Gemini, and human evaluation, respectively, indicating that the generated chains are largely aligned with the QA context while remaining grounded in the given evidence. The close agreement between LLM-based and human judgments further supports the reliability of the proposed evaluation.

% \begin{figure}[t]
% \centering
% \includegraphics[width=1\linewidth]{img/relevent&faithful.pdf}
% \caption{Evaluating Relevance and Faithfulness of Reasoning Chain.}
% \label{fig:Evaluation of Relevance and Faithfulness}
% \end{figure}

\section{Related Work}
\paragraph{Retrieval-Augmented Generation (RAG)} combines information retrieval to address issues like hallucination and outdated information \cite{gao2023retrieval,zhou-etal-2025-reflection,xu2024hallucination}. By integrating pre-retrieval (e.g., query rewriting), retrieval, post-retrieval (e.g., reranking), and generation \cite{fan2024survey,fang-etal-2025-karpa}, RAG \cite{sui-etal-2025-fidelis,zamani2024stochastic,yang2024rag,10.1145/3626772.3657660,dong2024don} improves the relevance of retrieved documents and enhances response reliability.

\paragraph{Large Language Models on Graph.} Graphs model relational data in various domains, and GNNs \cite{graphtransformer,gat,gcn} are effective for encoding graph structures. Recent works integrate LLMs with graph learning to enhance graph-based ACL tasks \cite{10.1145/3589335.3641251,jin2024large,jiang-etal-2025-hykge,ren2024survey}. Approaches range from feeding graph tokens into LLMs \cite{tang2024graphgpt,graphtoken,chai2023graphllm} to embedding graph neural layers within transformer architectures \cite{qin2023disentangled,zhu2024efficient,huang2024can,wang-etal-2025-knowledge-graph}.

\paragraph{Graph RAG.} Graph RAG incorporates graph-structured knowledge to enhance retrieval and reasoning \cite{baek2023knowledge,wang2024can,han2024retrieval,xu2024retrieval}. It utilizes graph databases  \cite{bollacker2008freebase,wikidata} to retrieve triples or subgraphs \cite{edge2024localgraphrag,mavromatis2024gnn,hu2024grag,wu2024stark}. Techniques like graph construction from text \cite{lightrag,fan2025minirag} and GNN-based encoding \cite{liu-etal-2025-ontology,jiang2022unikgqa,G-retriever,gnp} enhance reasoning. Recent approaches leverage LLMs for iterative reasoning over graphs \cite{Kg-agent,rog,tog,chen2024plan}. While Graph RAG emphasizes retrieval, it often lacks robust post-retrieval strategies. Our \ourmodel introduces a dual alignment mechanism to better link retriever and generator, improving subgraph-query alignment.

\section{Conclusion}
In this paper, we study GRAG and identify two key obstacles: (i) structure-coupled irrelevant evidence introduced by subgraph expansion and (ii) the structure-reasoning discrepancy between graph encoders and LLM semantics. To address these challenges, we propose \ourmodel, an anchor-and-rationale guided refinement framework. Experiments on three challenge benchmarks show consistent improvements over 18 strong baselines, demonstrating the effectiveness of the proposed approach. In addition, \ourmodel exhibits strong generalization across diverse GNN architectures and LLM backbones, indicating robust transferability to different encoder--reasoner pairings. Notably, the aligner’s structure-aware pruning substantially reduces input tokens and inference time while maintaining or even improving accuracy, highlighting \ourmodel as a practical refinement approach for efficient graph-grounded generation.

% \clearpage

\section*{Limitations} \label{sec:limitations}
Despite the promising results demonstrated by Align-GRAG, this work has certain limitations. Due to resource constraints, we were unable to conduct experiments on larger LLMs, leaving the effectiveness of the proposed alignment approach on more powerful models uncertain. Additionally, since our method requires the generation and utilization of graph embeddings, it cannot be directly implemented on closed-source models, which restricts access to internal embedding representations. These limitations highlight potential areas for future exploration, such as validating the scalability of the approach with state-of-the-art LLMs and developing techniques to adapt Align-GRAG for closed-source environments.

\section*{Ethical considerations}
Our method reduces hallucination by grounding generation in retrieved graph evidence, but it can still produce misleading answers if retrieval returns incomplete/noisy subgraphs or if extracted anchors/rationales are weakly grounded. Since the pipeline relies on LLMs for extraction and assessment, model biases and errors may propagate to both supervision and evaluation; therefore, outputs should be verified before use in high-stakes settings. What is more, we only use LLMs to polish the writing in this work.

% Bibliography entries for the entire Anthology, followed by custom entries
%\bibliography{anthology,custom}
% Custom bibliography entries only
\bibliography{custom}

% \clearpage
\appendix
% \newpage

% \lstdefinestyle{prompt}{
%     basicstyle=\ttfamily\fontsize{7pt}{8pt}\selectfont,
%     frame=none,
%     breaklines=true,
%     backgroundcolor=\color{lightgray},
%     breakatwhitespace=true,
%     breakindent=0pt,
%     escapeinside={(*@}{@*)},
%     numbers=none,
%     numbersep=5pt,
%     xleftmargin=5pt,
%     aboveskip=2pt,
%     belowskip=2pt,
% }
% \tcbset{
%   aibox/.style={
%     top=10pt,
%     colback=white,
%     % colframe=black,
%     % colbacktitle=black,
%     enhanced,
%     center,
%     % attach boxed title to top left={yshift=-0.1in,xshift=0.15in},
%     % boxed title style={boxrule=0pt,colframe=white,},
%   }
% }
% \newtcolorbox{AIbox}[2][]{aibox, title=#2,#1}

\section{Graph Retrieval}\label{sec:Graph Retriever}
Existing RAG methodologies are mainly designed for plain text documents or triplets, where information retrieval occurs independently of the graph structure \cite{fan2024survey,zhu2023large,pan2024unifying}. 
In the retrieval stage,  we first utilize an encoder-only language model (e.g., SentenceBERT \cite{sbert}) to encode textual information, including the query $ t_q $, text of each node $ t_n $ and edge $  t_e $ in the graph, respectively:
\begin{align}
    \boldsymbol{q} &= \text{SBERT}(t_q) \in \mathbb{R}^d, \\
    \boldsymbol{n} &= \text{SBERT}(t_n) \in \mathbb{R}^d, \\
    \boldsymbol{e} &= \text{SBERT}(t_e) \in \mathbb{R}^d
\end{align}

 Then we compute cosine similarity $\text{sim}(\cdot)$ between query embeddings $ \boldsymbol{q} $ and embeddings of nodes $\boldsymbol{n}$ and edges $\boldsymbol{e}$. 
 % and use top-$ k $ retrieval strategy to select the most relevant nodes and edges for the query:
 The top-$k$ nodes and edges are selected as the most relevant entities and relations.
\begin{align}
    \mathcal{V}_k &= \operatorname*{arg\,topk}_{\substack{\boldsymbol{n} \in \mathcal{V}}} \big( \text{sim}(\boldsymbol{q}, \boldsymbol{n}) \big), \\
    \mathcal{E}_k &= \operatorname*{arg\,topk}_{\substack{\boldsymbol{e} \in \mathcal{E}}} \big( \text{sim}(\boldsymbol{q}, \boldsymbol{e}) \big)
\end{align}
This forms top-$k$ entities and relation sets, denoted as $\mathcal{G}_{\text{retriever}}$. Inspired by \citet{G-retriever}, we further leverage the Prize-Collecting Steiner Tree algorithm (PCST) \cite{bienstock1993note} to maintain a controlled graph size.

To achieve the PCST algorithm, we assign ``prize'' to nodes and edges based on their similarity to a given query. Relevance is determined through the ranked sets of cosine similarity, $\mathcal{V}_k$ and $\mathcal{E}_k$, as follows:
\begin{equation}
\text{prize}(n) =
\begin{cases} 
    k - i, & \text{if } n \in \mathcal{V}_k \text{ and } n \\
        & \text{ is the } i\text{-th ranked node} \\
    0, & \text{otherwise}
\end{cases}
\end{equation}
where $ i $ is the rank of $ n $ in set $\mathcal{V}_k$. Nodes that are not among top $ k $ rankings are assigned a prize of zero. The objective of the PCST algorithm is to identify a subgraph that maximizes the total prize of nodes and edges while minimizing the cost:
\begin{align}
\small
\mathcal{G}_{\text{retriever}}
=
\underset{\substack{S \subseteq \mathcal{G},\\ S \text{ is connected}}}{\arg\max}&
\Bigg(
\sum_{n \in \mathcal{V}_S} \text{prize}(n)\\
+
\sum_{e \in \mathcal{E}_S} \text{prize}(e) \notag
&
-
\text{cost}(S)
\Bigg).
\end{align}

where $ \mathcal{V}_S $ and $ \mathcal{E}_S $ are the sets of nodes and edges in the subgraph $ S $, respectively. The cost of constructing the subgraph is defined as $\text{cost}(S) = |\mathcal{E}| \cdot C_e$,
where $ |\mathcal{E}| $ is the number of edges, and $ C_e $ is a predefined per-edge cost that serves as a regularization parameter to control the subgraph's size. In this way, we can obtain a preliminary retrieved subgraph.

\section{Experimental Details} \label{app:experimentaldetails}
\begin{table*}[htb]
\centering
\begin{tabular}{l|cccc}
\toprule
\textbf{Dataset} & \textbf{\#Training}  & \textbf{\#Validation} & \textbf{\#Test} & \textbf{Task Type} \\
\midrule 
 ExplaGraphs & 1,659& 553 & 554   &Commonsense reasoning\\
SceneGraphs& 59,978& 19,997  &20,025&Visual scene understanding\\
 WebQSP & 2,826 & 245 &1,628&Knowledge-based QA\\  
\bottomrule 
\end{tabular}
\caption{Statistics of GraphQA Benchmark \cite{G-retriever}.}
\label{statistics}
\end{table*}

\subsection{Datasets and Metrics.} \label{sec:appendix_dataset}
Following G-Retriever \cite{G-retriever}, we conducted experiments on the GraphQA benchmark \cite{G-retriever}, which includes ExplaGraphs (commonsense reasoning), SceneGraphs (scene graph understanding), and WebQSP (knowledge graph reasoning). The dataset statistics are summarized in Table \ref{statistics}, with a data split of train:validation:test = 6:2:2.
\begin{itemize}%[leftmargin=*]
\item \textbf{ExplaGraphs} is a dataset designed for commonsense reasoning and focuses on generating explanation graphs to predict stances in debates. It provides detailed, explicit, and commonsense-enriched graphs to evaluate arguments that either support or refute a given stance. The primary task is to determine whether an argument supports or opposes a stance.
\item \textbf{SceneGraphs} is a visual question answering dataset. Each graph provides detailed descriptions of objects, attributes, and relationships within an image. It is designed for tasks that involve spatial understanding and multi-step reasoning. The goal is to answer open-ended questions based on textual descriptions of scene graphs, with Accuracy used as the evaluation standard.
\item \textbf{WebQSP} is a large-scale, multi-hop knowledge graph question-answering dataset built on a subset of Freebase. It focuses on questions requiring multi-hop reasoning and includes facts within a two-hop neighborhood of the entities mentioned in the question. Since a single question may have multiple answers, the model's performance is evaluated using F1, Hit@1, and Accuracy following \citet{hu2024grag}.
\end{itemize}

\subsection{Baselines.}
To assess the effectiveness of our proposed method, we compare it against four categories of baselines:

\subsubsection{\textbf{Inference-only}.} 
This category includes approaches that employ frozen LLMs for question answering, using textual graph descriptions and queries as input. Different prompt strategies are considered, including:
\begin{itemize}%[leftmargin=*]
\item \textbf{Zero-shot}: directly answering questions based on retrieved information,
\item \textbf{Zero-CoT} \cite{cot}: enhances zero-shot reasoning by appending the phrase ``Let’s think step by step.'', which encourages the model to explicitly generate intermediate reasoning steps. 
\item \textbf{CoT-BAG} \cite{wang2024can}: adds ``Let’s construct a graph with the nodes and edges first.'' after providing the textual graph description, 
\item \textbf{KAPING} \cite{baek2023knowledge}: augments LLMs with knowledge graph facts at the prompt level, enabling more accurate QA without requiring model fine-tuning retrieving relevant graph triples.  
\item \textbf{ToG} \cite{tog}: integrates LLMs with KGs to support deep and reliable reasoning through interactive exploration of entities and relations.
\end{itemize}

\subsubsection{\textbf{Raw Fine-tuning.}} In this setting, following \cite{G-retriever}, we fine-tune the LLM via parameter-efficient methods, without applying advanced reranking techniques. Two widely used approaches are included:
\begin{itemize}%[leftmargin=*]
\item \textbf{Prompt Tuning} \cite{pt}: learns a set of continuous prompt embeddings that guide the frozen LLM during inference, enabling lightweight adaptation to specific tasks.  
\item \textbf{LoRA} \cite{lora}: introduces low-rank adaptation layers into transformer weights, achieving efficient fine-tuning with significantly reduced computational and memory costs.  
\end{itemize}

\subsubsection{\textbf{Reranker-based}.} 
These baselines employ reranking models to refine the ranking of candidate documents before passing them into the LLM. Representative methods include: 
\begin{itemize}%[leftmargin=*]
\item  \textbf{GTE} (General Textual Embedding) \cite{gte}: developed by Alibaba DAMO Academy, with two variants (gte-base, 109M parameters; gte-large, 335M parameters), trained on large-scale relevance pairs to improve retrieval accuracy, 
\item  \textbf{BGE} (BAAI General Embedding) \cite{bge}: cross-encoder models with two variants (bge-reranker-base, 278M parameters; bge-reranker-large, 560M parameters), optimized for retrieval-augmented generation, offering high accuracy at the cost of efficiency, 
\item  \textbf{G-RAG} and \textbf{G-RAG-RL} \cite{dong2024don}: reranker models that leverage GNNs to incorporate document connections and semantic cues from abstract meaning representation graphs for context-aware ranking.
\end{itemize}

\subsubsection{\textbf{GNN-based}.} 
This category integrates GNN encoders with LLM embeddings to enhance graph reasoning, including:
\begin{itemize}%[leftmargin=*]
\item  \textbf{GraphToken} \cite{graphtoken}: encodes graph structures as explicit prompts for LLMs, improving performance on graph reasoning tasks,  
\item  \textbf{GNP} \cite{gnp}: a plug-and-play framework that applies a GNN encoder with cross-modality pooling to enrich LLMs with knowledge graph information, 
\item  \textbf{G-Retriever} \cite{G-retriever}: introduces the GraphQA benchmark and employs retrieval-augmented generation with soft prompting, improving graph-based question answering, comprehension, and reducing hallucinations, 
\item  \textbf{GRAG} \cite{hu2024grag}: proposes a divide-and-conquer strategy for efficient textual subgraph retrieval and combines text and graph views within LLMs for graph context-aware generation.
\end{itemize}

\subsection{Implementation Details.} \label{app:implementation_details}
In this section, we further present the Implementation Details. To implement baselines, for GNP \cite{gnp}, we implemented the Graph Neural Prompting module within our framework, including components such as the GNN encoder, cross-modality pooling, and domain projector. For G-RAG and G-RAG-RL \cite{dong2024don}, we adopted their ranking approach, combining cross-entropy loss with pairwise ranking loss. However, since treating documents as nodes was infeasible in our case, we instead treated entities as nodes and employed the same GNN encoder as our method. We implemented GTE \cite{gte} and BGE-reranker \cite{bge} using their official open-source models. Nodes and triples are ranked by query similarity and, as in our method, fed into the LLM for generation. For the $\text{GRAG}_{PCST}$ model \cite{hu2024grag}, we reproduced its experiments. However, to ensure a fair evaluation, we standardized the retrieval process by using the PCST method for retrieval. This allowed us to directly compare it with their graph encoder approach.
Table~\ref{tab:hyperparamssetting} summarizes the hyperparameter settings for all datasets used in our experiments.

\begin{table}[t]
\centering
% \small
\begin{tabular}{l c}
\hline
\textbf{Hyperparameter} & \textbf{Value} \\
\hline
Number of Seed Nodes & 25 \\
Align Degree (Step) & 60 \\
Top-$K$ Retrieval & 10 \\
Batch Size & 8 \\
Max Epochs & 20 \\
Learning Rate & $1\times 10^{-5}$ \\
GNN Layers & 4 \\
GNN Hidden Dim  & 1024 \\
LoRA Rank  & 8 \\
LoRA Alpha  & 16 \\
LoRA Target Modules & \texttt{q\_proj, v\_proj} \\
\hline
\end{tabular}
\caption{Hyperparameter settings used in our experiments.}
\label{tab:hyperparamssetting}
\end{table}

\section{Impact of Hyperparameters}
In this section, we analyze \textbf{how hyperparameters affect performance}, including: the Number of Seed Nodes $n_\text{seed}$, Align Degree (training step) and Top K retrieval.

\subsection{\textbf{Impact of Seed Nodes}}\label{sec:Impact of Seed Nodes}
\begin{figure*}[t]
\centering
\includegraphics[width=0.32\linewidth]{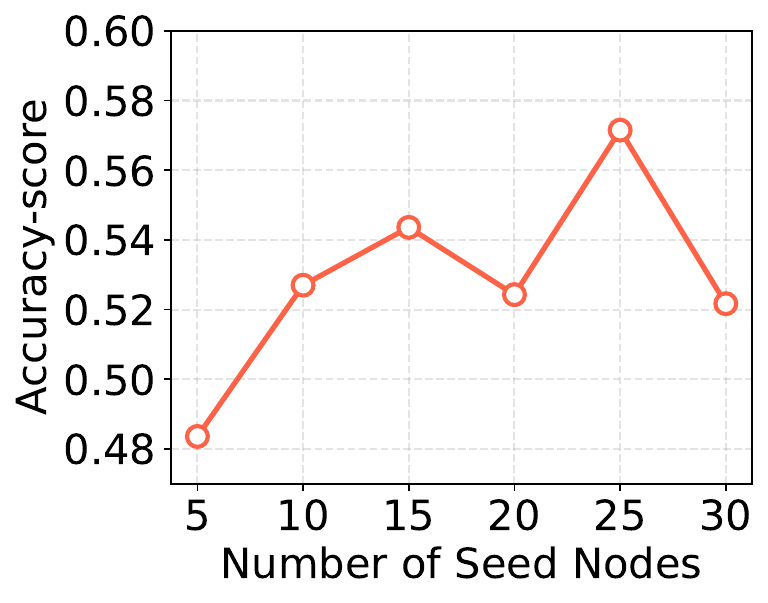}
\includegraphics[width=0.32\linewidth]{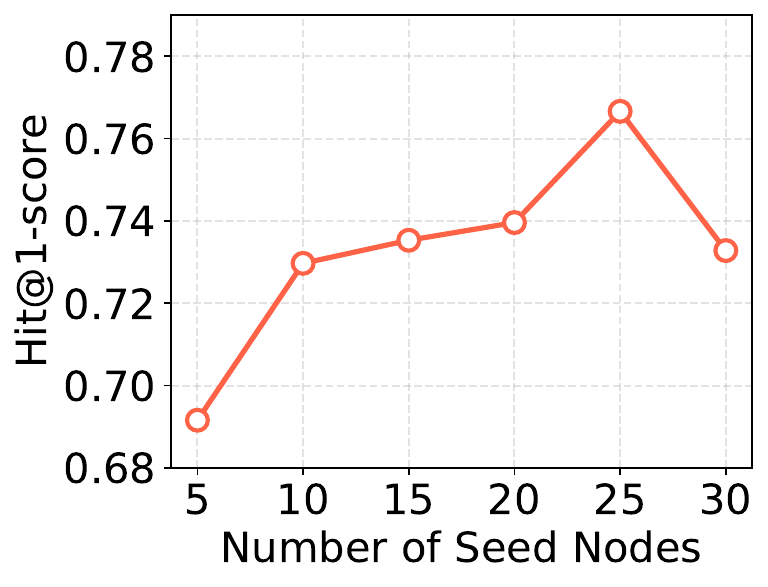}
\includegraphics[width=0.32\linewidth]{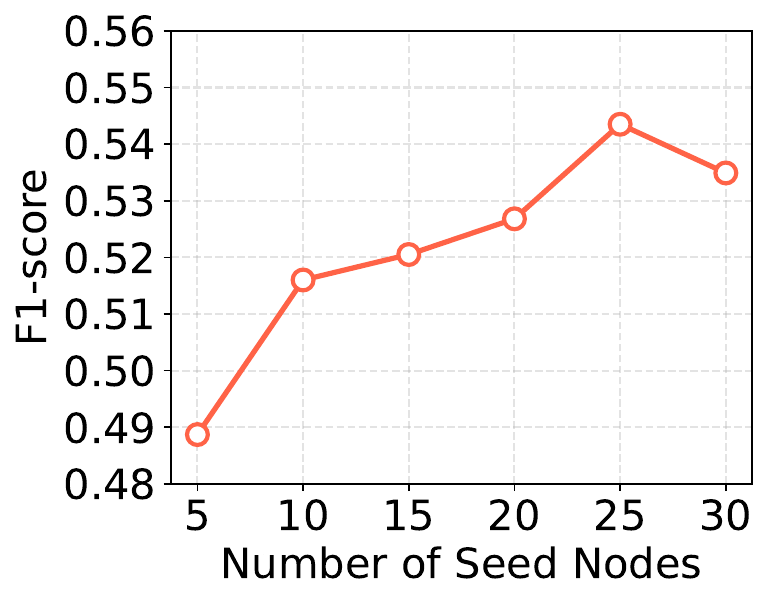}
\caption{Hyperparameters Analysis of the Number of seed nodes.}
\label{fig:Impact of Seed Nodes}
\end{figure*}

We analyze the impact of the \textbf{number of seed nodes} ($n_\text{seed}$) on model performance. The experiments are conducted on the WebQSP dataset, where we evaluate the Hit@1, F1 and Accuracy metrics. 
From the experiments in Figure \ref{fig:Impact of Seed Nodes}, we observe that the Hit@1, F1 and Accuracy performance peaks when the number of seed nodes is set to 25. Beyond this point (from 25 to 30), the performance starts to decline. This indicates that including too many nodes introduces a significant amount of irrelevant knowledge, which negatively impacts the model. Our pruning strategy effectively eliminates irrelevant knowledge to enhance model performance. 
On the other hand, when the number of seed nodes is as low as 5, the performance is considerably poor. This suggests that excessive pruning removes crucial knowledge, which is detrimental to performance. This highlights a trade-off: pruning reduces noise and improves performance, but over-pruning leads to the loss of essential knowledge.

\subsection{\textbf{Impact of Align Degree}}\label{sec:Impact of Align Degree} 

\begin{figure*}[t]
\centering
\includegraphics[width=0.32\linewidth]{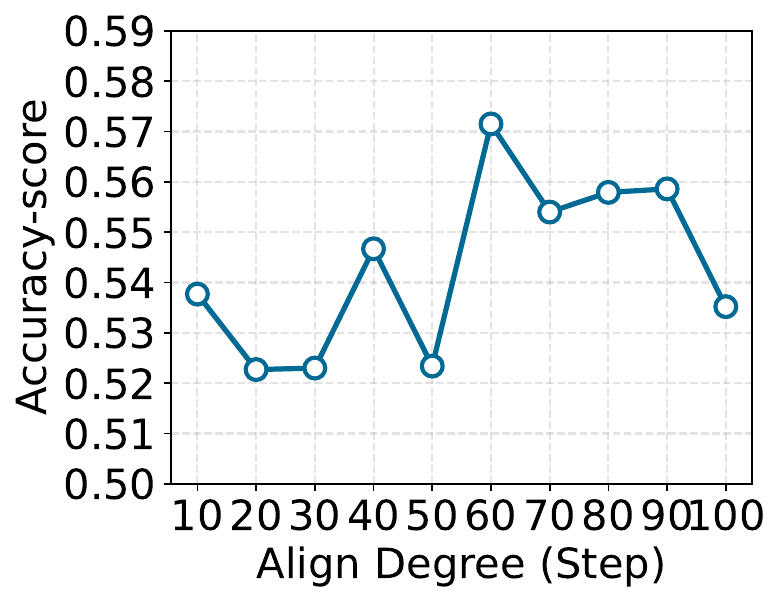}
\includegraphics[width=0.32\linewidth]{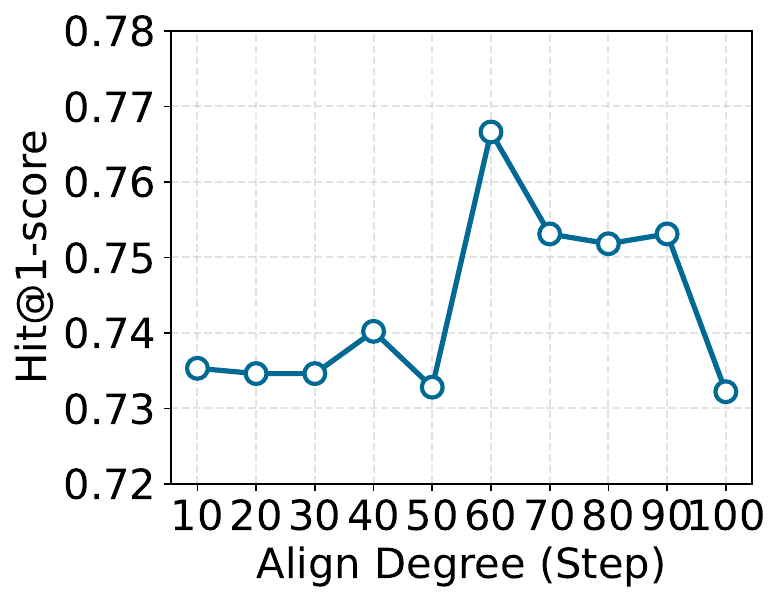}
\includegraphics[width=0.32\linewidth]{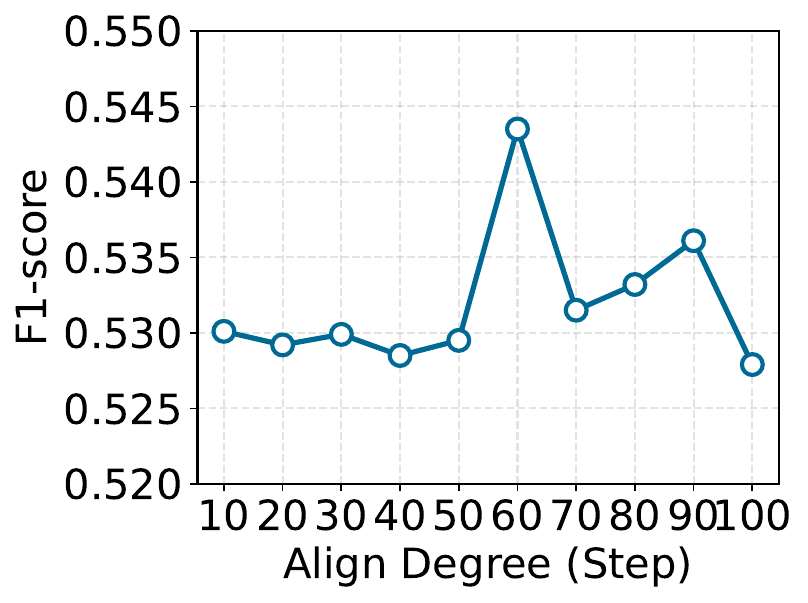}
\caption{Hyperparameters Analysis of Align degree.}
\label{fig:Impact of Align Degree}
\end{figure*}

This section examines how the \textbf{Align Degree} (number of training steps for Aligner module) influences model performance. As shown in Figure \ref{fig:Impact of Align Degree}, we evaluate the Hit@1, F1 and Accuracy metrics on the WebQSP dataset. 
From the experimental curve of Align Degree, we observe that the Hit@1, F1 and Accuracy metrics peak at around 60 epochs before declining. This indicates that, as training progresses, bridging the representation gap between the graph and language helps the LLM better understand graph data. 
However, excessive training may lead to overfitting, which disrupts graph information and ultimately causes a drop in performance. This suggests that there is an optimal point in training where the graph alignment is most effective, and training beyond this point can be detrimental.

\subsection{\textbf{Joint Impact of Seed Nodes and Align Degree}}\label{sec:HeatmapAnalysis}

\begin{figure*}[t]
\centering
\includegraphics[width=0.32\linewidth]{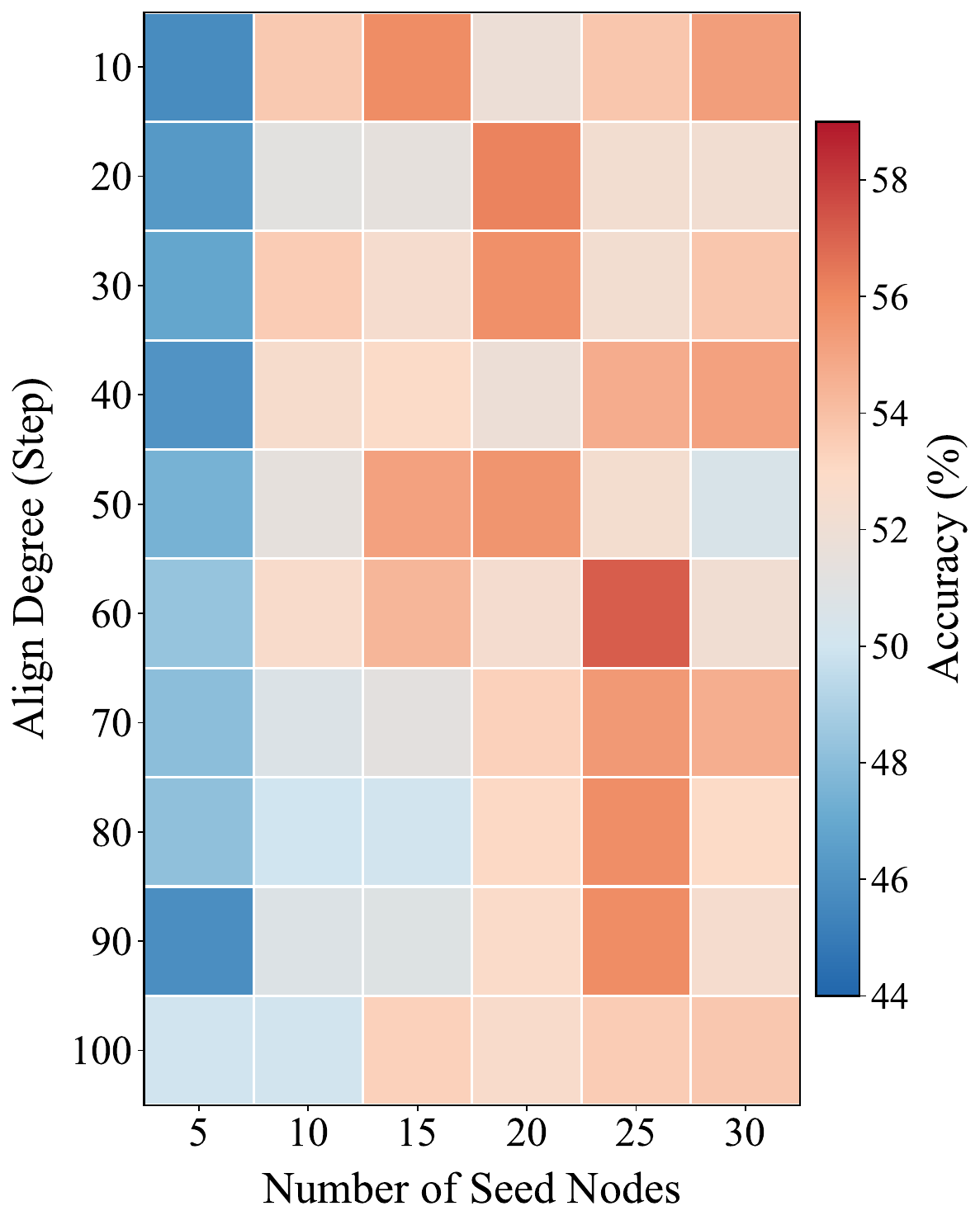}
\includegraphics[width=0.32\linewidth]{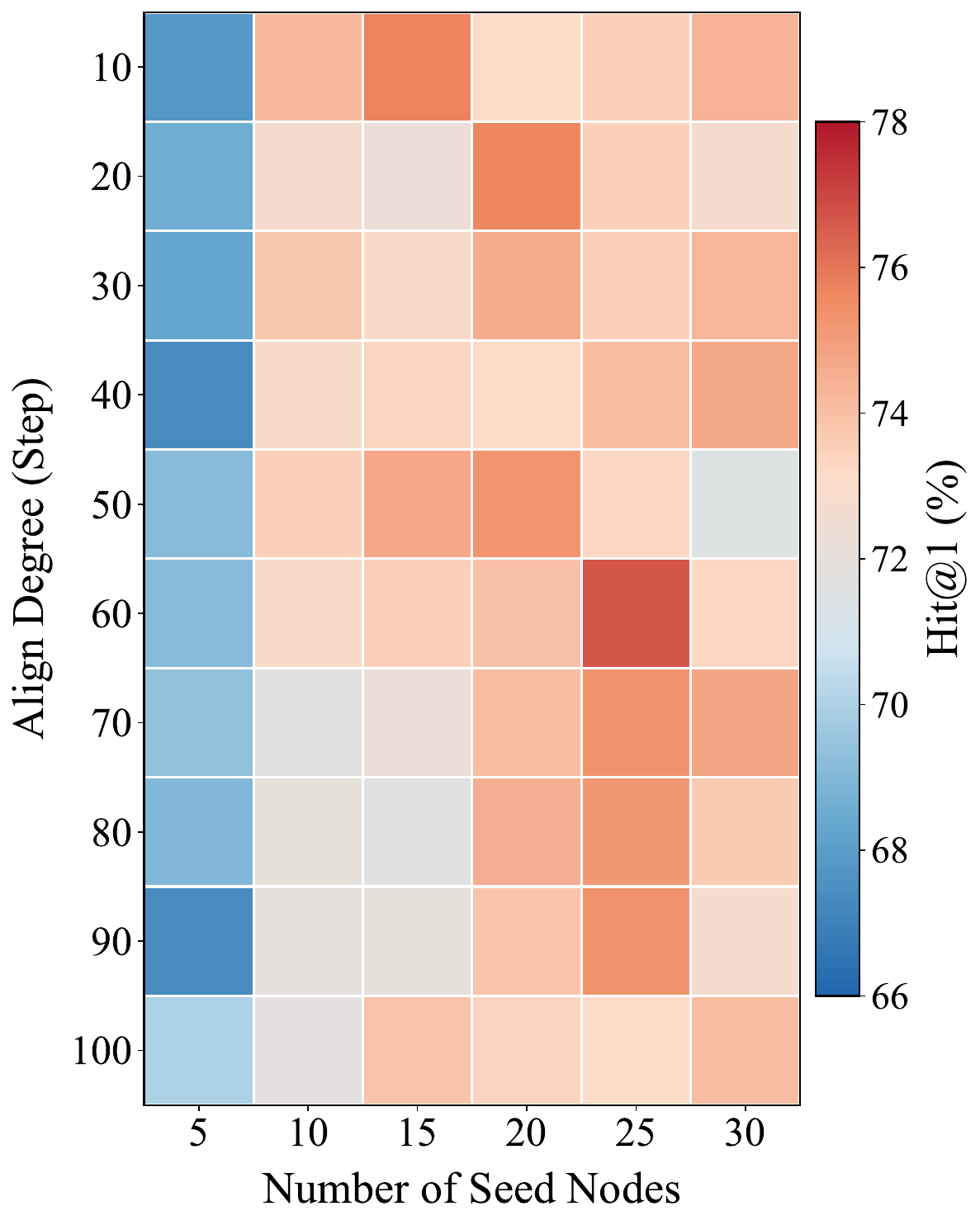}
\includegraphics[width=0.32\linewidth]{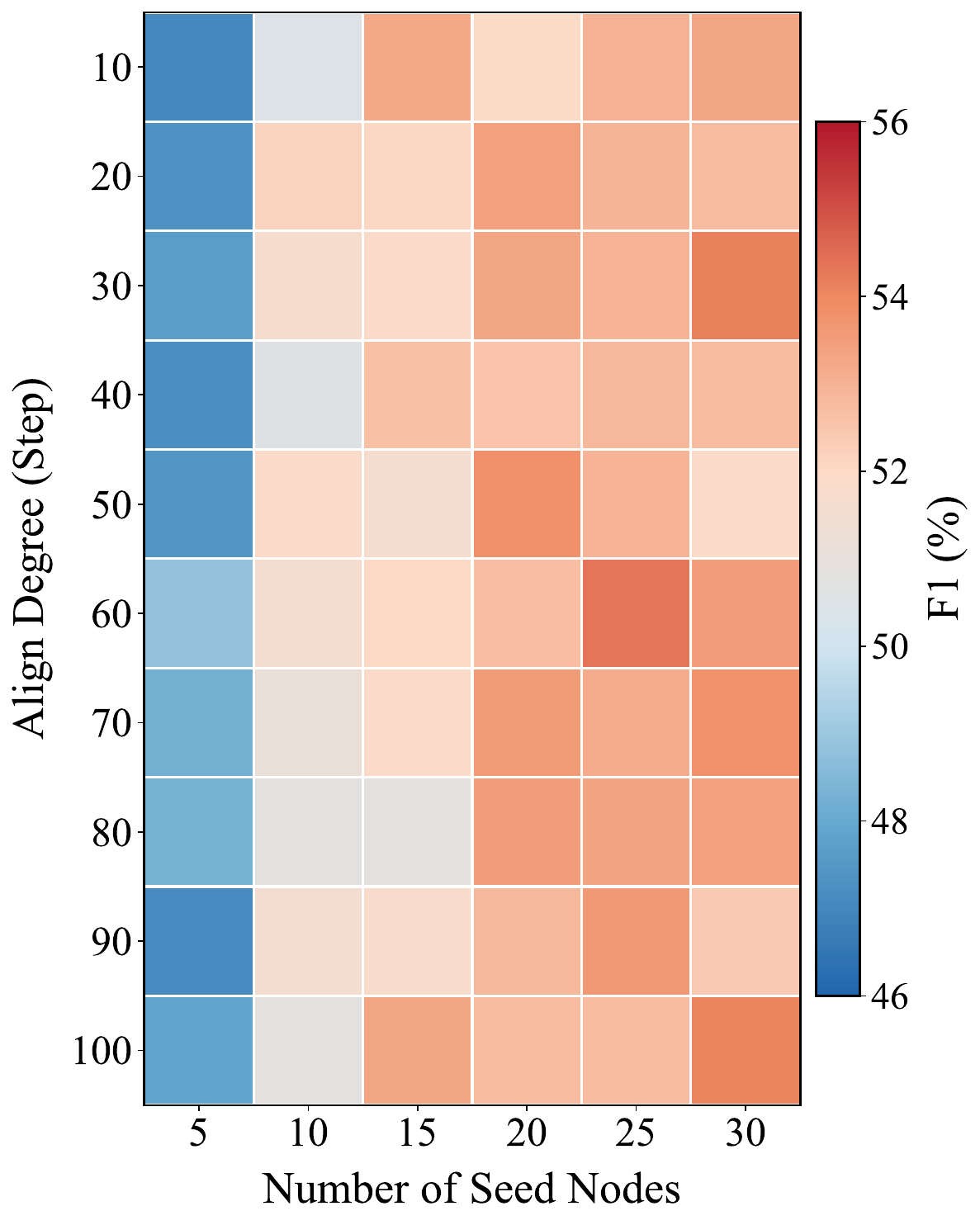}
\caption{Hyperparameter interaction heatmaps on WebQSP.}
\label{fig:Heatmap}
\end{figure*}

To further investigate the interaction between the \textbf{number of seed nodes} ($n_{\text{seed}}$) and the \textbf{Align Degree} (aligner training steps), we report a two-dimensional performance landscape in Figure~\ref{fig:Heatmap}. Each heatmap cell corresponds to one hyperparameter pair $(n_{\text{seed}}, \text{Align Degree})$, evaluated on WebQSP using Accuracy, Hit@1, and F1.
Overall, the three heatmaps exhibit a highly consistent trend: performance is maximized in a moderate regime, centered around $n_{\text{seed}}\!\approx\!25$ and Align Degree $\approx\!60$. When $n_{\text{seed}}$ is small (e.g., 5), all metrics remain low across nearly all Align Degrees, indicating that \textbf{insufficient retrieved evidence cannot be compensated} by longer aligner training. In contrast, when $n_{\text{seed}}$ becomes large (e.g., 30), the performance degrades, and the degradation is more pronounced under larger Align Degrees, suggesting that excessive retrieved noise may be amplified by over-training the aligner and can eventually harm the graph-language alignment quality. Similarly, for a fixed $n_{\text{seed}}$ in the mid-to-high range, increasing Align Degree improves results initially but starts to decline after the peak region, matching the \textbf{overfitting} pattern observed in the 1D analysis. These results confirm that the best performance requires balancing evidence coverage and noise control, where moderate $n_{\text{seed}}$ ensures adequate reasoning support while the aligner is trained sufficiently (but not excessively) to bridge the representation gap.

% \section{Hyperparameters Analysis}\label{Hyperparameters Analysis}
% In this section, we analyze \textbf{how hyperparameters affect performance}, including: the Number of Seed Nodes $n_\text{seed}$ and Align Degree (training step), as shown in Figure \ref{fig:hyperpara}. The experiments are conducted on the WebQSP dataset, where we evaluate the Hit@1 and Accuracy metrics.
% From experiments of $n_\text{seed}$, we observe that the Hit@1 and Accuracy performance peaks when the number of seed nodes is set to 25. Beyond this point (from 25 to 30), the performance starts to decline. This indicates that including too many nodes introduces a significant amount of irrelevant knowledge, which negatively impacts model. Our pruning strategy effectively eliminates irrelevant knowledge to enhance model performance. On the other hand, when the number of seed nodes is as low as 5, the performance is considerably poor. This suggests that excessive pruning removes crucial knowledge, which is detrimental. This highlights a trade-off: pruning reduces noise and improves performance, but over-pruning leads to the loss of essential knowledge.
% For the experimental curve of Align Degree, we observe that the Hit@1 and Accuracy metrics peak at around 60 epochs before declining. This indicates that, as training progresses, bridging the representation gap between the graph and language helps the LLM better understand graph data. However, excessive training may cause over-fitting, which disrupts the graph information, ultimately causing a drop in performance.

\subsection{\textbf{Impact of Top K retrieval}}\label{sec:Impact of Top K retrieval}
\begin{figure*}[t]
\centering
\includegraphics[width=0.32\linewidth]{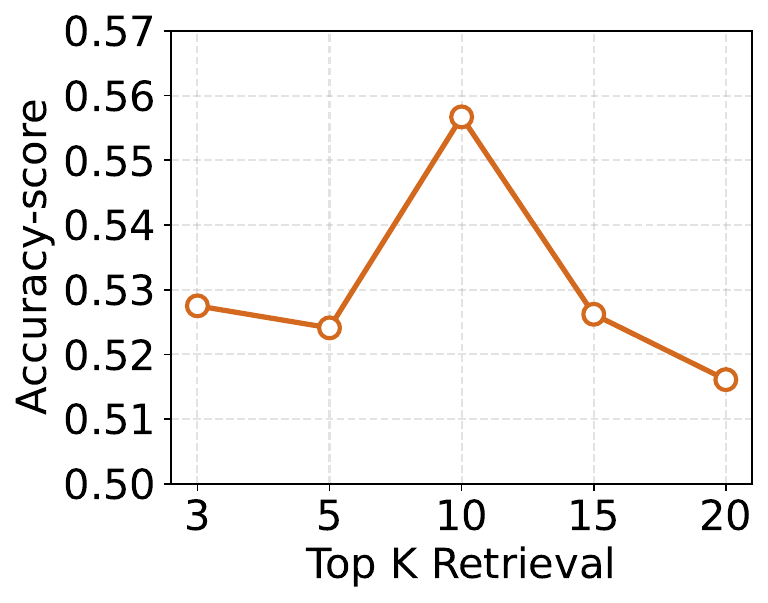}
\includegraphics[width=0.32\linewidth]{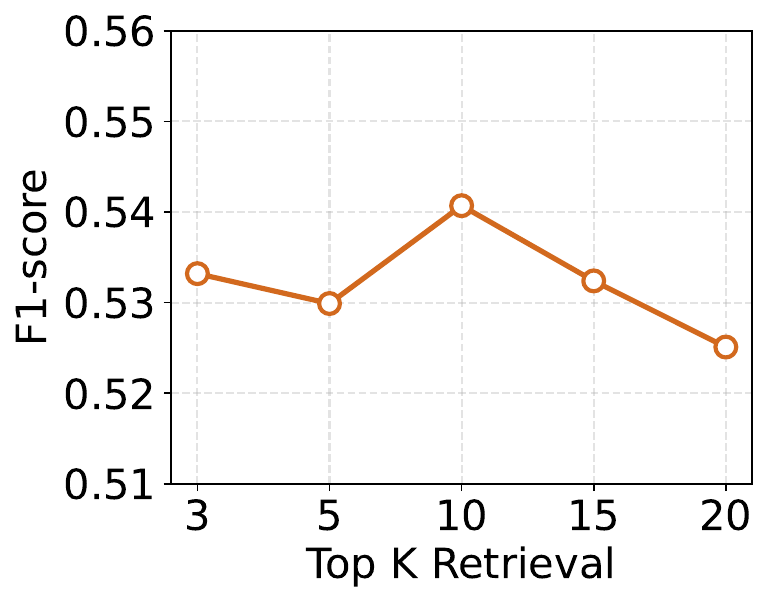}
\includegraphics[width=0.32\linewidth]{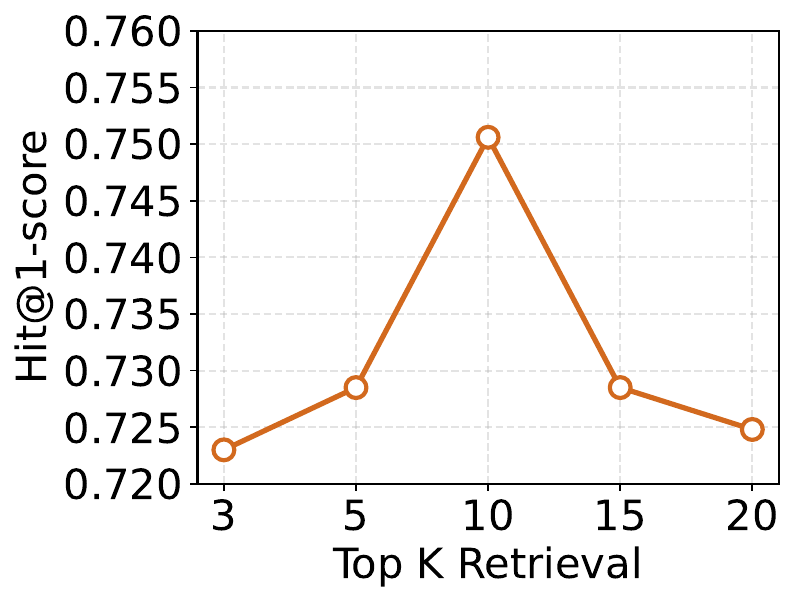}
\caption{Effect of Top K retrieval.}
\label{fig:hyperpara_topk}
\end{figure*}

Figure \ref{fig:hyperpara_topk} illustrates the impact of varying the \textbf{Top K retrieval} of entities and relations on model performance across Hit@1, F1 and Accuracy. By analyzing the trends in the graphs, we can derive insights into the effect of Top K on model performance. All three metrics show a similar trend: performance improves as K increases, peaks at K = 10, and then declines. This suggests that K = 10 strikes the optimal balance between retrieving relevant results and avoiding noise. Smaller K values may miss relevant information, while larger K values dilute relevance, reducing precision and retrieval quality. These findings highlight the importance of selecting an appropriate K value to maximize performance in retrieval-based systems.

% =========================
% Listings style (unchanged)
% =========================
\lstdefinestyle{prompt}{
    basicstyle=\ttfamily\fontsize{7pt}{8pt}\selectfont,
    frame=none,
    breaklines=true,
    backgroundcolor=\color{lightgray},
    breakatwhitespace=true,
    breakindent=0pt,
    escapeinside={(*@}{@*)},
    numbers=none,
    numbersep=5pt,
    xleftmargin=5pt,
    aboveskip=2pt,
    belowskip=2pt,
}

% =========================================================
% tcolorbox: keep your original AIbox style untouched
% and create a NEW box name + NEW color scheme for this case
% =========================================================
% \tcbset{
%   aibox/.style={
%     top=10pt,
%     colback=white,
%     % colframe=black,
%     % colbacktitle=black,
%     enhanced,
%     center,
%     % attach boxed title to top left={yshift=-0.1in,xshift=0.15in},
%     % boxed title style={boxrule=0pt,colframe=white,},
%   }
% }
% % (Optional) keep the original box if you still use it elsewhere
% \newtcolorbox{AIbox}[2][]{aibox, title=#2,#1}

% NEW: Only this box changes color, without affecting other boxes
\tcbset{
  casestudybox/.style={
    aibox,                       % inherit layout from aibox
    colback=blue!3,              % content background
    colframe=blue!55!black,      % border
    colbacktitle=blue!55!black,  % title background
    coltitle=white,              % title text
    boxrule=0.6pt,               % border thickness
    arc=2mm,                     % rounded corners (remove if unwanted)
    fonttitle=\bfseries,         % bold title
  }
}
% AIbox renamed to CaseStudyBox (use this one for the special colored box)
\newtcolorbox{CaseStudyBox}[2][]{casestudybox, title=#2,#1}

% ==== Minimal figure with ONLY text (no TikZ, no complex layout) ====
\begin{figure*}[!ht]
\centering
\small
\begin{CaseStudyBox}{Case Study of Extracted Rationale \& Anchors}
\vspace{1mm}

\textbf{Extraction\_Prompt}\\
\textbf{Question}: what is the name of the first harry potter novel?\\
\textbf{Answer}: harry potter and the philosopher's stone\\
\hrule
\vspace{1mm}

\textbf{Graph DataBase}: \\
\small
\begin{verbatim}
node_id,node_attr
0,harry potter and the chamber of secrets
1,harry potter and the philosopher's stone
3,j. k. rowling
7,harry potter and the half-blood prince
9,harry potter and the prisoner of azkaban
11,harry potter and the goblet of fire
16,harry potter
24,harry potter and the deathly hallows
57,fiction
59,harry potter literary series
76,professor severus snape
98,fantasy
......
src,edge_attr,dst
16,freebase.equivalent_topic.equivalent_domain,91
91,freebase.domain_profile.featured_views,806
199,book.written_work.subjects,455
59,book.book_subject.works,199
3,book.author.works_written,670
24,media_common.adapted_work.adaptations,46
59,book.book_subject.works,371
0,book.book.characters,325
24,book.book.genre,57
24,book.book_edition.book,24
9,book.book.genre,98
190,book.book_edition.book,24
1,book.book.genre,224
24,book.book_edition.book,24
7,book.book.genre,98
59,book.literary_series.fictional_universe,16
478,book.book_edition.book,24
......
\end{verbatim}
\normalsize

\hrule
\vspace{1mm}

\textbf{Extracted Rationale}: \\
\begin{itemize}[leftmargin=1.2em, topsep=0pt, itemsep=1pt]
    \item The question asks for the name of the \textbf{first harry potter novel}, and the graph includes a specific book node labeled \textbf{harry potter and the philosopher's stone}.
    \item The graph’s provided summary states that \textbf{j. k. rowling} wrote the first Harry Potter novel, explicitly naming it as \textbf{harry potter and the philosopher's stone}.
    \item Therefore, using the graph evidence, the first Harry Potter novel is \textbf{harry potter and the philosopher's stone}, matching the given answer.
\end{itemize}

\textbf{Extracted Anchors}: \\
\begin{itemize}[leftmargin=1.2em, topsep=0pt, itemsep=1pt]
    \item \textbf{entity}: \textbf{harry potter and the philosopher's stone}
    \item \textbf{entity}: \textbf{j. k. rowling}
    \item \textbf{relation}: \textbf{book.author.works\_written}
\end{itemize}

\vspace{1mm}
\end{CaseStudyBox}
\caption{The case for extracting the anchors and rationales. The extracted rationale chain identifies the critical intermediate node \texttt{j. k. rowling} and links it to the answer \texttt{harry potter and the philosopher's stone}.}
\label{fig:summarization_text}
\end{figure*}

\section{Case study}\label{sec:case study}
In this case study, we investigate how summarization enhances the discovery of critical intermediate nodes within anchors and rationales, as illustrated in Table \ ref{tab:summarization}. Using well-crafted prompts in conjunction with a graph database, we show that concise, targeted summaries can surface pivotal entities—such as `J.K. Rowling'—that bridge the semantic gap between an initial query and the final answer. Our analysis evaluates the extent to which summarization reliably extracts these key connectors, thereby enabling coherent multi-hop reasoning that would otherwise be obscured within the raw graph structure. We further demonstrate that, without the interpretive layer provided by summarization, the underlying graph database lacks the necessary contextual signals to assemble the correct reasoning path end-to-end, underscoring the value of summarization as a catalyst for interpretable and effective knowledge retrieval.

\section{LLM Prompts}
We introduce three prompts: the Extraction Prompt, the Generator Prompt, and the Relevance and Faithfulness Prompt. Each serves a specific function in enhancing the performance and evaluation of LLMs. Below is a detailed explanation of their purposes and applications, as illustrated in Figure \ref{fig: Summarization Prompt}, Figure \ref{fig: Generator Prompt}, and Figure \ref{fig: Relevance and Faithfulness Prompt}.

\begin{figure*}[!ht] 
\begin{AIbox}{Anchor and Rationale Extraction Prompt}
\vspace{1mm}
You are a helpful assistant for \textbf{anchor and rationale extraction} from graph evidence.

\textbf{Inputs}: a question, its answer, and a textualized graph database (nodes/edges described in text).
\textbf{Goal}: extract (1) a \emph{rationale chain} that links the question to the answer using graph evidence, and (2) a small set of \emph{anchors} (key intermediate entities/relations) that can be grounded to graph nodes/edges.

\textbf{Definitions}:
\begin{itemize}[leftmargin=1.2em, topsep=0pt, itemsep=1pt]
    \item \textbf{Rationale chain}: an ordered list of short reasoning steps. Each step should mention the evidence used and move closer to the answer.
    \item \textbf{Anchors}: reasoning-critical intermediate entities/relations that appear in the graph database and are necessary for the reasoning.
\end{itemize}

\textbf{Constraints}:
\begin{itemize}[leftmargin=1.2em, topsep=0pt, itemsep=1pt]
    \item Use only information supported by the provided graph database and the given answer.
    \item Anchors must be \textbf{verbatim spans} from the graph database (copy exact surface forms).
    \item Keep each rationale step concise.
\end{itemize}

\textbf{Output format} (follow exactly):
\begin{enumerate}[leftmargin=1.2em, topsep=0pt, itemsep=1pt]
    \item \textbf{RationaleChain}: a numbered list of steps (3--6 steps).
    \item \textbf{Anchors}: a bullet list of anchors. For each anchor, provide its type (\texttt{entity} or \texttt{relation}) and the copied span.
\end{enumerate}

\textbf{Question}: \{question\}

\textbf{Answer}: \{Answer\}

\textbf{Graph DataBase}: \{Textualized Graph\}

\textbf{Now produce the output.}
\end{AIbox}
\vspace{-1em}
\caption{Prompt for anchor and rationale extraction.}
\label{fig: Summarization Prompt}
\end{figure*}

\textbf{The Extraction Prompt} is specifically designed to generate a comprehensive and cohesive anchor and rationale by synthesizing three key inputs: a given question, its corresponding answer, and the relevant textualized graph data. The goal of this prompt is to create a logical connection between the question and the answer, while also integrating information extracted from the provided graph data. This ensures the output not only answers the question but also demonstrates an understanding of the contextual relationships within the graph. By combining these elements, the prompt facilitates the generation of a rich, informative summary that serves as a bridge between structured and unstructured data.

\begin{figure*}[t] 
\begin{AIbox}{Relevance and Faithfulness Prompt}
\vspace{1mm}
\textbf{Evaluation of Relevance:}

Evaluate the relevance of the anchor and rationale in answering the QUESTION. The relevant anchor and rationale contain information that helps answer the question, even if partially. Return one of the following labels: 'Relevant', or 'Irrelevant' without any additional response.
\vspace{1mm}

\textbf{Evaluation of Faithfulness:}

Evaluate the following anchor and rationale for faithfulness in answering the QUESTION. A faithful response should include information that helps answer the question, even if partially, avoid inventing new details, and not contradict the context. Return one of the following labels: 'Faithful' or 'Not Faithful' without any additional response.

\end{AIbox}
\vspace{-1em}
\caption{Evaluation of Relevance and Faithfulness Prompt.}
\label{fig: Relevance and Faithfulness Prompt}
\end{figure*}

\begin{figure*}[t] 
\begin{AIbox}{Generator Prompt}
\vspace{1mm}
\textbf{Prompt for WebQSP and SceneGraphs datasets:}

Textualized Graph: \{Textualized Graph\}.

Please answer the given question.
Question: \{question\}

Answer:

\vspace{1mm}
\textbf{Prompt for ExplaGraphs dataset:}

Textualized Graph: \{Textualized Graph\}.

Argument 1: \{arg1\}

Argument 2: \{arg1\}

Question: Do argument 1 and argument 2 support or counter each other? Answer in one word in the form of `support' or `counter'. 

Answer:
\end{AIbox}
\vspace{-1em}
\caption{Prompt for Generator.}
\label{fig: Generator Prompt}
\end{figure*}

\textbf{The Relevance and Faithfulness Prompt} is an evaluation-focused tool that serves two distinct but complementary purposes:
\begin{itemize}
    \item Relevance Evaluation: This component assesses whether the anchor and rationale provided is relevant to the given question. A relevant anchor and rationale must contain information that contributes meaningfully to answering the question, even if the contribution is partial. The model is required to return one of two labels: Relevant or Irrelevant, with no additional commentary.
    \item Faithfulness Evaluation: This component measures the faithfulness of anchor and rationale in responding to the question, which should accurately represent the information in the context, avoid fabrications, and not contradict the provided data. Similarly, the model outputs one of two labels: Faithful or Not Faithful. This dual evaluation ensures the generated responses are both accurate and contextually grounded.
\end{itemize}

\textbf{The Generator Prompt} is tailored to the dataset being used, with its structure varying depending on the specific requirements of the dataset. For WebQSP and SceneGraphs, the prompt focuses on answering a question by leveraging the provided textualized graph data. This encourages the model to extract and synthesize relevant information from graph representations to deliver a concise and accurate response. For ExplaGraphs, the prompt shifts its focus to reasoning. It requires the model to evaluate the relationship between two arguments (e.g., whether they support or counter each other), based on the provided textualized graph data. The model outputs a single-word response, either support or counter, reflecting the nature of the relationship.

\end{document}